\documentclass[runningheads]{llncs}

\usepackage[mobile]{eccv}

\usepackage{eccvabbrv}

\usepackage{graphicx}
\usepackage{booktabs}

\usepackage[accsupp]{axessibility}  

\usepackage{overpic}
\usepackage{bm}
\usepackage{pifont}
\usepackage{multirow}
\usepackage[ruled,linesnumbered,vlined]{algorithm2e}

\usepackage{colortbl,xcolor,soul}

\definecolor{LightCyan}{rgb}{0.88,1,1}
\definecolor{mp}{RGB}{112,173,71}
\definecolor{iijgr}{rgb}{1,0,0}

\newcommand{\myPara}[1]{\vspace{.02in}\noindent\textbf{#1}}

\usepackage{floatpag}

\usepackage{subcaption}

\usepackage[misc]{ifsym} 

\makeatletter
\def\@fnsymbol#1{\ensuremath{\ifcase#1\or \dagger\or \ddagger\or
   \mathsection\or \mathparagraph\or \|\or **\or \dagger\dagger
   \or \ddagger\ddagger \else\@ctrerr\fi}}
\makeatother

\usepackage{wrapfig}
\usepackage{makecell}

\definecolor{cvprblue}{rgb}{0.21,0.49,0.74}

\usepackage{orcidlink}

\begin{document}

\title{MAGR: Manifold-Aligned Graph Regularization\\for Continual Action Quality Assessment}

\titlerunning{MAGR: Manifold-Aligned Graph Regularization for CAQA}

\author{
Kanglei Zhou\inst{1,3}
\and
Liyuan Wang\inst{2} \and
Xingxing Zhang\inst{2}\and
Hubert P. H. Shum\inst{3}\and
Frederick W. B. Li\inst{3}\and
Jianguo Li\inst{4}\and
Xiaohui Liang\inst{1,5}%
\textsuperscript{(\Letter)}%
}

\authorrunning{K. Zhou et al.}

\institute{
State Key Laboratory of VR Technology and Systems, Beihang University  
\and
Department of Computer Science and Technology, Institute for AI, BNRist Center, Tsinghua-Bosch Joint ML Center, THBI Lab, Tsinghua University
\and
Department of Computer Science, Durham University
\and
Children's Hospital of Capital Institute of Pediatrics
\and
Zhongguancun Laboratory
\\ \email{liang\_xiaohui@buaa.edu.cn}
}

\maketitle

\vspace{-0.4cm}
\begin{abstract}
Action Quality Assessment (AQA) evaluates diverse skills but models struggle with non-stationary data. We propose Continual AQA (CAQA) to refine models using sparse new data.
Feature replay preserves memory without storing raw inputs. 
However, the misalignment between static old features and the dynamically changing feature manifold causes severe catastrophic forgetting. 
To address this novel problem, we propose Manifold-Aligned Graph Regularization (MAGR), which first aligns deviated old features to the current feature manifold, ensuring representation consistency. It then constructs a graph jointly arranging old and new features aligned with quality scores.
Experiments show MAGR outperforms recent strong baselines with up to 6.56\%, 5.66\%, 15.64\%, and 9.05\% correlation gains on the MTL-AQA, FineDiving, UNLV-Dive, and JDM-MSA split datasets, respectively. This validates MAGR for continual assessment challenges arising from non-stationary skill variations.
Code is available at \href{https://github.com/ZhouKanglei/MAGR_CAQA}{https://github.com/ZhouKanglei/MAGR\_CAQA}.
\end{abstract}

\vspace{-0.5cm}
\section{Introduction}
Action Quality Assessment (AQA) evaluates performance beyond recognition~\cite{zhou2023hierarchical}. Traditional AQA methods trained on small static datasets~\cite{parmar2019and,parmar2019action,zhou2024cofinal} struggle with dynamically changing skills in sports~\cite{xu2022finediving,zhang2023logo} and rehabilitation~\cite{zhou2023video,yao2023contrastive,liao2020deep}, requiring updated standards. Continual Learning (CL) provides promising solutions to non-stationarity~\cite{zhang2023few,wang2023comprehensive}, yet faces challenges like catastrophic forgetting in sequential training~\cite{wang2023comprehensive,wang2023incorporating}. By enabling continuous learning with memory preservation, CL facilitates lifelong adaptation and stability.

\begin{figure}[htp]
    \centering
    \includegraphics[width=0.8\linewidth,clip,trim=85 120 80 110]{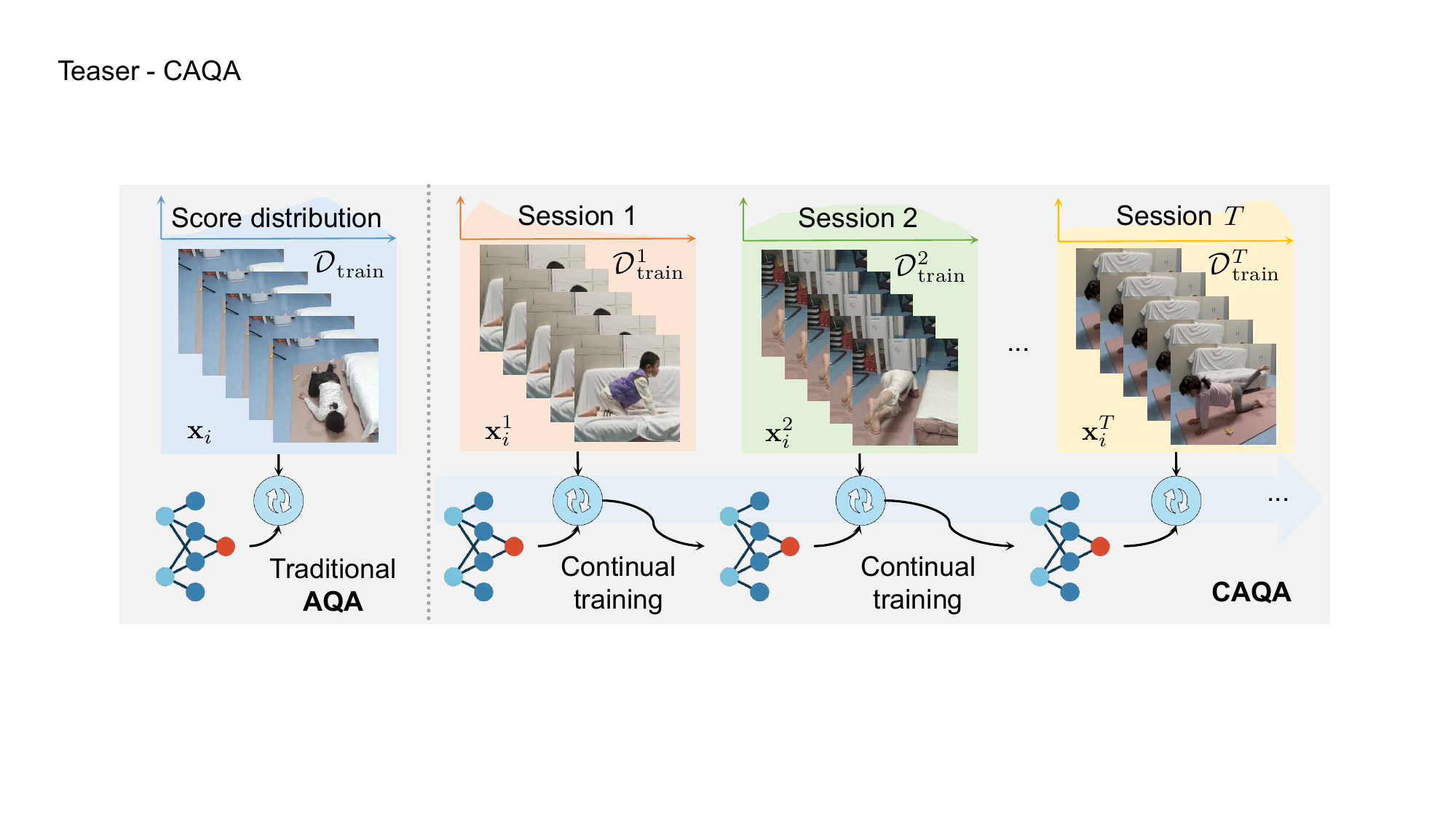}
    \caption{
        Traditional AQA vs CAQA: 
        CAQA refines AQA from a few sequentially arrived instances without exhaustive retraining, which advances CL beyond classification.
    }
    \label{fig:caqa}
\end{figure}

While CL offers promising solutions for dynamic skill assessments, prior work has scarcely explored its application to AQA, which poses a unique CL challenge through its reliance on subtle quality regression over evolving feature manifolds, defining a novel problem of continual assessment. To address this issue, we introduce Continual AQA (CAQA) to seamlessly refine AQA models using sparse new data (see \cref{fig:caqa}) without catastrophic forgetting. Unlike traditional CL research focused on discrete classifications \cite{chi2022metafscil,wang2023hierarchical,wang2021afec}, CAQA involves continuous quality score regression requiring adaptation to changing quality score distributions characterizing how skills evolve over time. To advance CAQA research, we propose the novel task of incrementally refining AQA models over sequential sessions using only a few arriving samples, fulfilling real-world needs while posing distinct challenges compared to traditional classification tasks in CL. 

A key challenge in CAQA implementation is the misalignment between static old features and the evolving feature manifold, resulting in catastrophic forgetting. Effectively retaining and utilizing past knowledge is essential across sequential sessions. While experience replay methods like those in \cite{rolnick2019experience,kukleva2021generalized} are common in CL, they raise privacy concerns, especially in sensitive AQA domains like sports training or medical care. Feature replay methods, such as those proposed in \cite{yang2023neural,zhang2023slca}, offer privacy advantages but struggle with adapting to new data distributions. The dilemma of using a fixed backbone limits model adaptability, while updating the backbone risks severe catastrophic forgetting by misaligning old features with the evolving feature manifold. Addressing this specific misalignment issue and updating backbones are critical for CAQA's adaptability.


To mitigate the misalignment, we propose Manifold-Aligned Graph Regularization (MAGR), a novel feature replay method consisting of two essential steps. Firstly, MAGR iteratively learns the manifold shift between sessions using only the current session's raw data due to privacy constraints, ensuring accurate alignment of old features to the current feature manifold. Secondly, MAGR focuses on readjusting the feature distribution with the quality score distribution from local and global perspectives to eliminate confusion of features across sessions and within the same session, thereby improving assessment accuracy.
%
%

To stimulate research in this important yet understudied field, we establish a comprehensive CAQA benchmark study. This includes splitting multiple AQA datasets, defining custom evaluation protocol and metrics, and incorporating recent strong baselines.
Experiments demonstrate that MAGR outperforms recent strong baselines by a substantial margin, achieving correlation gains of up to 6.56\%, 5.66\%, 15.64\%, and 9.05\% on the MTL-AQA, FineDiving, UNLV-Dive, and JDM-MSA split datasets, respectively.
Our main contributions are:

\begin{itemize}
    \vspace{-0.1cm}
    \item To the best of our knowledge, we are the first to introduce CAQA to enable efficient AQA model refinement using sparse new data, addressing the unique challenges versus traditional classification tasks in CL.

    \item To address the misalignment, we propose MAGR as a novel solution, aligning old features to the current manifold without raw inputs and ensuring alignment between feature and quality score distributions.

    \item We validate MAGR on multiple AQA split datasets, demonstrating superior performance over recent strong baselines and establishing its effectiveness for continual performance assessment, thereby advancing CL and AQA research.
\end{itemize}
\section{Related Work}


\myPara{Action Quality Assessment (AQA)} evaluates the quantitative performance of performed actions in various areas \cite{wang2021survey}, such as sports \cite{zhou2023hierarchical}, medical rehabilitation \cite{zhou2023video,ding2023sedskill}, and skill assessment \cite{smith-etal-2020-put,wang2020towards}. Earlier methods depended heavily on hand-crafted features and heuristics, revealing certain limitations. 
By integrating deep learning, various models \cite{parmar2019and,tang2020uncertainty,yu2021group} have shown improved performance.
Due to label scarcity, existing AQA datasets \cite{wang2021survey} are relatively small, risking over-fitting. To mitigate this, pre-trained backbones such as C3D \cite{tran2015learning}, I3D \cite{carreira2017quo}, and VST \cite{liu2022video} are commonly employed. 
As every frame may contain essential AQA cues \cite{parmar2017learning}, videos are often segmented into clips for separate processing due to the limited computation resources, which hinders a complete understanding of the action. To address this, a GCN-based method \cite{zhou2023hierarchical} has been proposed to eliminate semantic ambiguities. 
Despite the success, the ever-evolving nature of individual skills and environments requires lifelong adaptation in AQA. To address this practical problem, we seamlessly incorporate CL paradigms into AQA systems to ensure efficient adaptation.

\myPara{Continual Learning (CL)}, also known as incremental or lifelong learning \cite{wang2023comprehensive}, aims to train a model over a sequence of tasks, ensuring the retention of learned tasks and mitigating catastrophic forgetting. 
Since AQA is significantly restricted by relatively small datasets, we primarily focus on few-shot CL \cite{zhang2023few,zhao2023few}.
which suffers from severe over-fitting. Recent strategies to combat this issue include preserving representation topology \cite{tao2020few}, constructing exemplar relation graphs for knowledge distillation \cite{dong2021few}, 
and using generative replays of earlier data distributions \cite{liu2022few}. 
These advanced CL strategies often rely on the experience replay of old samples \cite{wang2021memory}, raising legitimate privacy concerns \cite{zhang2023few}. This is particularly critical in sensitive AQA domains such as proprietary sports training and medical rehabilitation.
Feature replay methods \cite{yang2023neural,zhang2023slca,wang2023hierarchical} offer privacy benefits and have demonstrated strong performance. However, generative replay \cite{wang2021ordisco,wang2021triple} faces limitations due to the variable quality of the generated data, which can impact its effectiveness in real-world scenarios. 
Moreover, a fixed backbone \cite{yang2023neural} simplifies the model but limits its adaptability to evolving real-world scenarios, while an updated backbone \cite{zhang2023slca} risks misalignment with old features, deteriorating the forgetting issue. 
Our work is dedicated to addressing the misalignment with backbone updates in the context of CAQA.
\vspace{-0.2cm}
\section{Continual Action Quality Assessment (CAQA)}




We propose the novel task of CAQA to adapt evolving individual skills or health conditions over time. For instance, in athlete rehabilitation, movement quality evolves with recovery stages, challenging traditional AQA systems trained on static datasets. Unlike traditional CL methods focused on classification \cite{wang2023comprehensive,zhao2023few}, CAQA involves continuous quality score regression, which is crucial for accurately capturing subtle variations in performance. This aspect is especially relevant in real-world scenarios where subtle variations in quality assessment can have significant implications, presenting a unique and pressing challenge in the field. By introducing CAQA, we aim to address this critical gap and provide a robust framework for continual refinement of AQA models in dynamic environments.
The following defines AQA and CAQA. 

\myPara{AQA Task} aims to assign a quality score $\hat{y}\in\mathcal{Y}$ (typically $y\in\mathbb{R}$) to an input video $\mathbf{x}\in\mathcal{X}$ (typically $\mathbb{R}^{L\times H\times W\times 3}$), where $L$, $H$, $W$ and 3 represent length, height, width, and channels of inputs, respectively. The goal is to learn mappings $\hat{y} = g_{\theta_g}(\bm{h})$ and $\bm{h} = f_{\theta_f}(\mathbf{x})$ between $\mathcal{X}$ and $\mathcal{Y}$ from data $\mathcal{D}_{\mathrm{train}} = \{(\mathbf{x}_n, y_n)\}_{n=1}^N$ using encoders $f(\cdot)$ and regressors $g(\cdot)$, where $\bm{h}$ denotes the latent feature. 

\myPara{CAQA Task} aims to seamlessly integrate CL into AQA to enable continuous adaptation of assessment capabilities. It processes sequentially obtained datasets $\{\mathcal{D}_{\mathrm{train}}^t\}_{t=1}^T$ over $T$ sessions. A key challenge is catastrophic forgetting when learning new sessions. To address this, CAQA employs feature replay utilizing a memory bank $\mathcal{M}^{t-1}$ to store old features. The objective is formulated as:
\begin{equation} \label{eq_caqa}
\min_{\theta_f, \theta_g} ~ \mathcal{L}_{\mathrm{D}} + \mathcal{L}_{\mathrm{M}},
\end{equation} 
where $\mathcal{L}_{\mathrm{D}}$ and $\mathcal{L}_{\mathrm{M}}$ are score regression losses on current data $\mathcal{D}_{\mathrm{train}}^t$ and memory bank $\mathcal{M}^t$, enabling incremental refinement without exhaustive retraining.

\section{Manifold-Aligned Graph Regularization (MAGR)}
We first motivate MAGR and then explain its major novel technical components.

\begin{figure}
    \centering
    \includegraphics[width=0.5\linewidth,clip,trim=140 276 140 76]{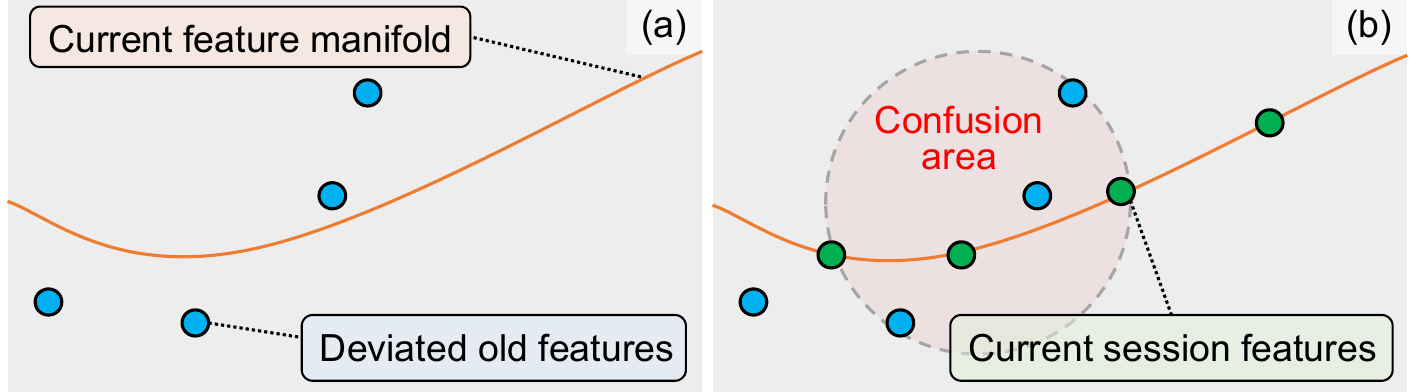}\includegraphics[width=0.5\linewidth,clip,trim=140 80 140 270]{figs/idea.pdf}
    \caption{
        Our core idea: (a) Deviation of old features (blue circles) from the current manifold (orange curve) caused by the manifold shift; (b) Potential confusion for score regression due to the mixture of old features and current session features (green circles); (c) Translation of old features from the previous manifold (yellow curve) to the current; (d) Readjustment of the feature distribution to align with the quality score distribution.
    }{
    \phantomsubcaption\label{fig:problem-a}%
    \phantomsubcaption\label{fig:problem-b}%
    \phantomsubcaption\label{fig:problem-c}%
    \phantomsubcaption\label{fig:problem-d}%
     }%
    \label{fig:problem}
    \vspace{-0.1cm}
\end{figure}






\subsection{Motivation and Pipeline of MAGR}
\myPara{Motivation.}
We propose MAGR as a solution to address the key challenge of CAQA. Its motivation is threefold: (1) To mitigate the catastrophic forgetting, we adopt feature replay rather than raw data replay to prioritize user privacy which is crucial for sensitive AQA domains; (2) To improve the adaptability, the complexity of AQA requires backbone updating that induces the misalignment between static old features and dynamically evolving feature manifolds (see \cref{fig:problem-a,fig:problem-b}); and (3) To tackle the misalignment, MAGR adopts a two-step alignment process by dynamically translating deviated features (see \cref{fig:problem-c}) and aligning the feature distribution (see \cref{fig:problem-d}).


\begin{figure}
    \centering
    \includegraphics[width=\linewidth,clip,trim=145 335 130 215]{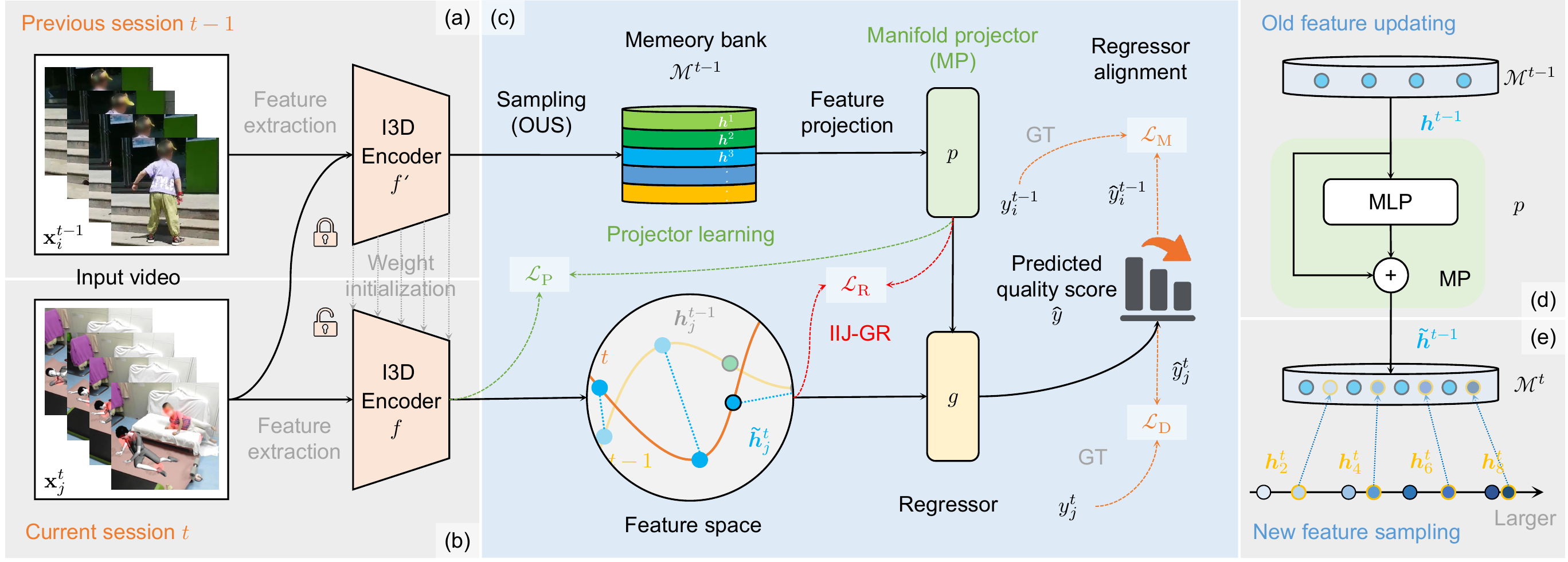}
    \caption{
    MAGR framework: (a) At the end of session $t-1$, representative samples are chosen and stored in the memory bank $\mathcal{M}^{t-1}$, and the feature extractor $f'$ is frozen. (b) Throughout session $t$, {\color{mp} MP} translates old features to the current manifold, while {\color{iijgr} IIJ-GR} regulates the entire feature space to align with the quality space. (c) After that, old features are first updated. (d) Then, new features are selected for the updated memory bank, denoted as $\mathcal{M}^{t}$, where the superscript indicates the update session.
    }{
    \phantomsubcaption\label{fig:framework-a}%
    \phantomsubcaption\label{fig:framework-b}%
    \phantomsubcaption\label{fig:framework-c}%
    \phantomsubcaption\label{fig:framework-d}%
    \phantomsubcaption\label{fig:framework-e}%
    }%
    \label{fig:framework}
\end{figure}

\myPara{Pipeline and Insight.}
\cref{fig:framework} depicts the MAGR framework, where we consider two consecutive sessions $t-1$ and $t$. 
At the end of session $t-1$, we employ Ordered Uniform sampling (OUS) to select representative features that are then stored in a memory bank $\mathcal{M}^{t-1}$ to log the old distribution.   
OUS involves sorting the entire training set before performing uniform sampling, ensuring coverage across the whole range. 
In comparison to DER \cite{buzzega2020dark}, OUS maximally preserves the old quality score distribution, thereby improving CAQA performance (refer to \cref{tab:ablation_study_mtl}).
During session $t$, one branch is dedicated to refining AQA by learning new data, while the other retrieves a mini-batch of old features from the memory bank $\mathcal{M}^{t-1}$ to maintain the memory stability. 
Old features are refined using the Manifold Projector (MP, see \cref{sect_manifold_projector}), while the Intra-Inter-Joint Regularizer (IIJ-GR, see \cref{sect_graph_reg}) aligns both old and new features with the quality score distribution. Together, the objective in \cref{eq_caqa} can be reformulated as:
\begin{equation} \label{eq_opt}
\begin{aligned}
    \min_{\Theta} ~ & \mathcal{L}_{\mathrm{D}} + \mathcal{L}_{\mathrm{M}} + \lambda_{\mathrm{P}} 
 \mathcal{L}_{\mathrm{P}} + \lambda_{\mathrm{R}} \mathcal{L}_{\mathrm{R}},
    \\ \text{s.t.} ~~ & \hat{y}^s_i = g_{\theta_g}(p_{\theta_p}(\bm{h}^s_i)), ~ (\bm{h}_i^s, y_i^s) \in \mathcal{M}^t,
    \\ & \hat{y}_j^t = g_{\theta_g}(f_{\theta_f}({\mathbf{x}}^t_j)), ~(\mathbf{x}_j^t, y_j^t) \in \mathcal{D}_{\mathrm{train}}^t, 
\end{aligned}
\end{equation}
where $\bm{h}_i^s =  f_{\theta_f}(\mathbf{x}^s_i)$ denotes old features, $\Theta=\{\theta_f, \theta_p, \theta_g\}$ is the parameter set, and $\mathcal{L}_{\mathrm{P}}$ and $\mathcal{L}_{\mathrm{R}}$ encourage correcting deviated features and regulating the feature space, respectively. $\lambda_{\mathrm{P}}$ and $\lambda_{\mathrm{R}}$ balance the two constraints.
In addition, the training process is detailed in \cref{sect_algorithm}.


\subsection{Manifold Projector: Deviated Feature Translation} 
\label{sect_manifold_projector}

\begin{figure}
    \centering
    \begin{minipage}{0.62\textwidth}
        \includegraphics[width=\linewidth,clip,trim=100 250 160 60]{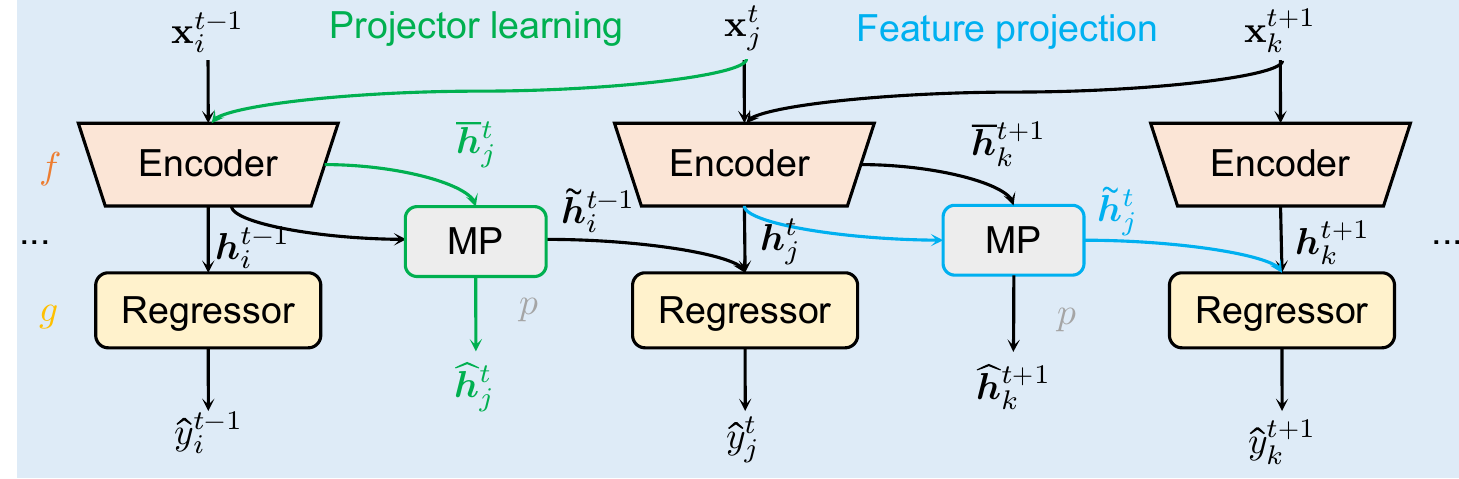}
    \end{minipage}
    \begin{minipage}{0.37\textwidth}
    \caption{Illustrations of MP: 
    Projector learning estimates the manifold shift, and feature projection translates old features to the current manifold.}
        \label{fig:mp}
    \end{minipage}
\end{figure}

MP is designed to learn a mapping from the previous manifold to the current one.
\cref{fig:mp} highlights two sub-steps of MP across three consecutive sessions: projector learning and feature projection.


\myPara{Projector Learning} is designed to estimate the manifold shift at each model update using dependencies from the current session data.
We replicate and freeze the encoder $f'(\cdot)$ from the previous session at the beginning of session $t$. This allows us to obtain the initial feature $\bm{\bar{h}}_j^t$ for the current session data $\mathbf{x}_j^t$ as a base to estimate manifold shifts. For simplicity, we will omit arguments in the lower-right corners of functions in subsequent discussions. 
Next, $\bm{\bar{h}}_j^t$ is passed through the projector $p$ to obtain the predicted updated feature $\bm{\hat{h}}_j^t$, which is:
\begin{equation} \label{eq_mp_learning}
\bm{\hat{h}}^t_j = \bm{\bar{h}}_j^t + p(\bm{\bar{h}}_j^t) 
,~\text{where}~\bm{\bar{h}}_j^t = f'(\mathbf{x}_j^t).
\end{equation}
Here, we observed that the inclusion of a residual link in feature projection enhances effectiveness in learning through updates (refer to the results in \cref{tab:ablation_study_mtl}).

The learning process of MP is supervised by minimizing the difference between the predicted and actual updated features of the current session data:
\begin{equation} \label{eq_lp}
\mathcal{L}_{\mathrm{P}} = \frac{1}{|\mathcal{D}_{\mathrm{train}}^t|} \sum_j \| \bm{h}^t_j - \bm{\hat{h}}^t_j \|{}_2^2,
\end{equation}
where $|\cdot|$ indicates the set size, and $\bm{h}_j^t = f(\mathbf{x}_j^t)$ is the actual updated feature.


\myPara{Feature Projection}
is designed to translate old features to the current manifold.
For each old feature ${\tilde{\bm{h}}}^{s}_i~(i=1,2,\cdots,|\mathcal{M}^{t-1}|)$ from the memory bank, its corrected feature can be calculated by:
\begin{equation} \label{eq_fp}
\tilde{\bm{h}}^{s}_i = \tilde{\bm{h}}_i^{s} + p(\tilde{\bm{h}}_i^{s}) 
,~\text{where}~s=1,2,\cdots,t-1.
\end{equation}

\myPara{Benefit of MP.}
By leveraging dependencies from the current session to correct deviated old features, MP plays a crucial role in addressing the manifold shift in feature-replay methods \cite{zhang2023slca,yang2023neural}. Its novelty is its ability to adaptively learn the manifold shift between sessions without needing access to raw old inputs. This sets it apart from traditional experience replay methods involving raw data \cite{rolnick2019experience}. MP offers both privacy and resource-efficiency advantages, making it particularly valuable in sensitive AQA domains. In addition, MP overcomes the restriction of fixing backbones found in \cite{yang2023neural}, providing high adaptability to real-world complexities. Utilizing an MLP with residual links, MP presents a simple yet effective solution for addressing misalignment in feature replay.

\subsection{\small Intra-Inter-Joint Graph Regularizer: Feature Distribution Alignment} \label{sect_graph_reg}

While MP has translated deviated old features with the current feature manifold, it may not ensure alignment between the feature distribution and its quality score distribution. In \cref{fig:problem-b}, it is difficult to discern the relative relationships of quality scores between inter-session features (across different sessions) and intra-session features (within the same session), posing challenges for quality score regression and ultimately impacting AQA performance. To this end, we propose IIJ-GR to regulate the feature space for accurate score regression.

\begin{figure}
    \centering
    \begin{minipage}{0.62\textwidth}
        \includegraphics[width=\linewidth,clip,trim=95 165 120 140]{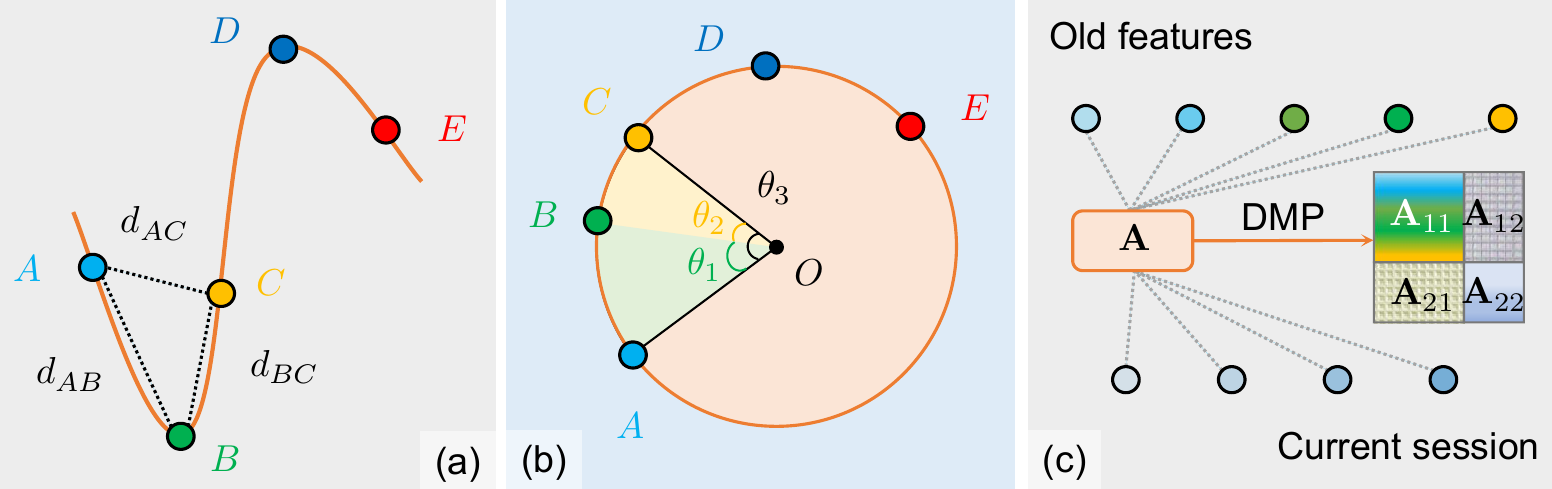}
    \end{minipage}
    \begin{minipage}{0.37\textwidth}
        \caption{Illustrations of IIJ-GR: (a) Euclidean distance, (b) Angular distance, and (c) Distance Matrix Partitioning (DMP).}
        {
        \phantomsubcaption\label{fig:iij_gr-a}%
        \phantomsubcaption\label{fig:iij_gr-b}%
        \phantomsubcaption\label{fig:iij_gr-c}%
        }
        \label{fig:iij_gr}
    \end{minipage}
\end{figure}

\cref{fig:iij_gr} depicts the fundamental concept behind our proposed IIJ-GR module. 
We identify a significant limitation in graph-based CL methods \cite{tao2020few,dong2021few}, where Euclidean distance is commonly utilized to measure point separation within a manifold. Indeed, the quality score relationship inherent in the data particularly meets the geodesic property. However, using the Euclidean distance often fails to satisfy such relationship $d_{AC} = d_{AB} + d_{BC}$ (see \cref{fig:iij_gr-a}). To address this problem, we leverage the insight that geodesic distance is proportional to angular differences \cite{bao2023pcfgaze}. Consequently, we normalize features to fit a unit sphere and utilize these angular differences for distance estimation (see \cref{fig:iij_gr-b}). Additionally, to effectively regulate the feature space, we propose partitioning the distance matrix (see \cref{fig:iij_gr-c}). This matrix encapsulates dependencies from both previous and current sessions. Our proposed Distance Matrix Partitioning (DMP) first divides the matrix into block matrices, which are then collaboratively optimized to ensure a quality-aware manifold from both local and global perspectives.

\myPara{Distance Matrix Partitioning.} 
Let's consider a mini-batch $b_1$ from previous sessions, symbolized by $\bm{h}_i^s$, and a batch $b_2$ from the current session, represented by $\bm{h}_j^t$. Concatenating them results in the feature matrix $\mathbf{H} = [\bm{h}_1^s, \bm{h}_2^s, \cdots, \bm{h}_{b_1}^s, \bm{h}_1^t, \allowbreak \bm{h}_2^t, \cdots, \bm{h}_{b_2}^t] \in \mathbb{R}^{(b_1+b_2) \times D}$. Similarly, we can obtain the corresponding score vector $\bm{y}\in \mathbb{R}^{(b_1+b_2) \times 1}$. In this way, the distance matrix $\mathbf{A}\in \mathbb{R}^{(b_1+b_2)\times(b_1+b_2)}$ can be calculated as:
\begin{equation}
\mathbf{A} = \arccos({\tilde{\mathbf{H}}\tilde{\mathbf{H}}^\top}),\text{where}~\tilde{\mathbf{H}} = \mathbf{H} / \|\mathbf{H}\|.
\end{equation}
Thereafter, $\mathbf{A}$ can be divided into four sub-matrices: $\mathbf{A}_{11}\in \mathbb{R}^{b_1\times b_1}$ captures relationships within the previous sessions, $\mathbf{A}_{12}\in \mathbb{R}^{b_1\times b_2}$ and $\mathbf{A}_{21}\in \mathbb{R}^{b_2\times b_1}$ characterize relationships between the previous and current sessions, $\mathbf{A}_{22}\in \mathbb{R}^{b_2\times b_2}$ encapsulates relationships within the current session, and $\mathbf{A}$ as a whole signifies the integrated relationships spanning all observed sessions.

\myPara{Graph Regularization.}
Essentially, this step involves creating a quality score distance matrix $\mathbf{S} = \bm{y}-\bm{y}^\top$, which acts as supervision for $\mathbf{A}$. To achieve this, we first define a loss function to gauge the distribution discrepancy between two matrices $\mathbf{P},\mathbf{Q} \in \mathbb{R}^{N\times N}$ as:
\begin{equation}
L(\mathbf{P}, \mathbf{Q}) = \frac{1}{N} \sum_{i=1}^N \mathrm{KL}({\sigma}(\mathbf{P}_i), {\sigma}(\mathbf{Q}_i)),
\end{equation}
where $\mathrm{KL}(\cdot)$ represents the Kullback–Leibler (KL) divergence, and ${\sigma}(\cdot)$ denotes the softmax function.
Here, the KL divergence offers a looser and more holistic constraint compared to MSE, aligning well with the correlation evaluation metric and ultimately leading to better performance (refer to \cref{tab:ablation_study_mtl}). 
Then, the total regularization loss can be represented as:
\begin{equation} \label{eq_lr}
\mathcal{L}_{\mathrm{R}} = L(\mathbf{A}, \mathbf{S}) + \sum_{i=1}^2\sum_{j=1}^2 L(\mathbf{A}_{ij}, \mathbf{S}_{ij}),
\end{equation}
where the partition of $\mathbf{S}$ is the same as that of $\mathbf{A}$.
This aids in closely aligning the underlying feature space with the actual quality distribution of the data. 

\myPara{Benefit of IIJ-GR.}
IIJ-GR offers several advantages over manual pair selection in contrastive loss \cite{bai2022action}. It effectively captures complex feature relationships from both local and global perspectives, which helps mitigate catastrophic forgetting and ensures assessment improvement across different sessions. 
While kernel alignment \cite{cortes2012algorithms} aims to maximize the alignment between kernels in a static feature space, IIJ-GR explicitly learns to align the raw feature space itself with the quality score space in a dynamic manner. Compared to graph-based CL methods \cite{tao2020few,dong2021few}, DMP considers both inter-session and intra-session constraints, enhancing feature representation fidelity and aligning it with the quality space.

\subsection{Training Procedure} \label{sect_algorithm}
In \cref{algo_procedure}, the training begins with careful parameter initialization. The encoder leverages pre-trained weights from \cite{zhou2023hierarchical}, providing domain knowledge, while other components are randomly initialized to adapt to CAQA sessions.

For each session, we maintain a previous encoder copy to compute the relative manifold shift for learning the MP. During training, for each batch, we extract features using the encoder and predict scores through a regressor. If memory is non-empty, we employ feature replay to mitigate catastrophic forgetting. MAGR incorporates MP and IIJ-GR for feature space alignment. We then replay a mini-batch of these corrected features. Model parameter optimization is iterative until convergence. At the end of the session, old features are first updated, and then representative features are drawn and stored in the memory bank.


\begin{algorithm}
    \small
    \renewcommand{\algorithmcfname}{\small Algorithm}
    \caption{\small The training process of MAGR.}
    \label{algo_procedure}
    \KwIn{Training datasets $\mathcal{D}_{\mathrm{train}}^t, t\in \{1,2,\cdots,T\}$, and the network parameter $\Theta = \{\theta_{f}, \theta_{g}, \theta_p\}$.}
    \KwOut{The trained model with the optimal parameter $\Theta$.}
    Initialize $\theta_{f}$ with the pre-trained I3D weight, randomly initialize $\theta_{p}$ and $\theta_{g}$, and initialize  $\mathcal{M}$ with $\varnothing$\; 
    \For{$t \gets 1, 2, \cdots, T$}{
        $f' \gets \mathrm{copy}(f)$ \tcp*{copy and fix the previous encoder}
        \While{not converged}{
            $\hat{y}_i^{t} \gets g(\bm{h}_i^t),~ \bm{h}_i^{t} \gets f(\mathbf{x}_i^t)$ \tcp*{$i=1,2,\cdots, b_2$}
            $\mathcal{L}_{\mathrm{D}} \gets 1/b_2 \sum_i (\hat{y}_i^t - y_i^t)^2$ \tcp*{ $(\mathbf{x}_i^t, y_i^t) \in \mathcal{D}_{\mathrm{train}}^t$}
            \If{$\mathcal{M}^t \neq \varnothing$}{
                $\bm{\hat{h}}^t_i \gets f'(\mathbf{x}_i^t) + p(f'(\mathbf{x}_i^t))$\tcp*{feature projection in \cref{eq_mp_learning}}
                $\mathcal{L}_{\mathrm{P}} \gets 1/b_2\sum_i^{b_2} \|\bm{h}^t_i - \bm{\hat{h}}^t_i\|_2^2$ \tcp*{feature learning in \cref{eq_lp}}
                Calculate $\mathcal{L}_{\mathrm{R}}$ using \cref{eq_lr}\tcp*{IIJ-GR in \cref{sect_graph_reg}}
                $\tilde{\bm{h}}^s_i \gets \tilde{\bm{h}}^s_i + p(\tilde{\bm{h}}^s_i)$\tcp*{$s<t$, \cref{eq_fp}}
                $\mathcal{L}_{\mathrm{M}} \gets 1/b_1 \sum_i^{b_1} (\hat{y}_i^s - y_i^s)^2,~ \hat{y}_i^s \gets g(\tilde{\bm{h}}^s_i)$\tcp*{regressor alignment}
            }
            Update $\Theta$ by optimizing \cref{eq_opt}\tcp*{backward}
        }
        Update old features $\tilde{\bm{h}}^{s}$ from $\mathcal{M}^{t-1}$ to $\mathcal{M}^{t}$\;
        Draw representative features $\bm{h}^{t}$ to $\mathcal{M}^{t}$ using the OUS strategy\;
    }
\end{algorithm}

\section{The CAQA Benchmark}
We construct a comprehensive benchmark study for advancing CAQA research.

\subsection{Datasets}
We choose four datasets, namely MTL-AQA \cite{parmar2019action}, FineDiving \cite{xu2022finediving}, UNLV-Dive \cite{parmar2017learning}, and MSA-JDM \cite{zhou2023video}, to ensure a holistic evaluation across diverse domains and scenarios, each with varying sample sizes. Leveraging representative AQA datasets spanning sports and medical care domains allows us to address privacy concerns and ensure the generalization of CAQA models. For more details about each dataset, please refer to the supplementary material.

\subsection{Experiment Protocol}
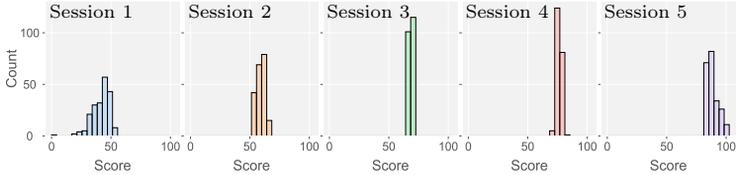
\begin{figure}
    \centering
        \begin{overpic}[width=0.8\linewidth,clip,trim=0 0 0 13]{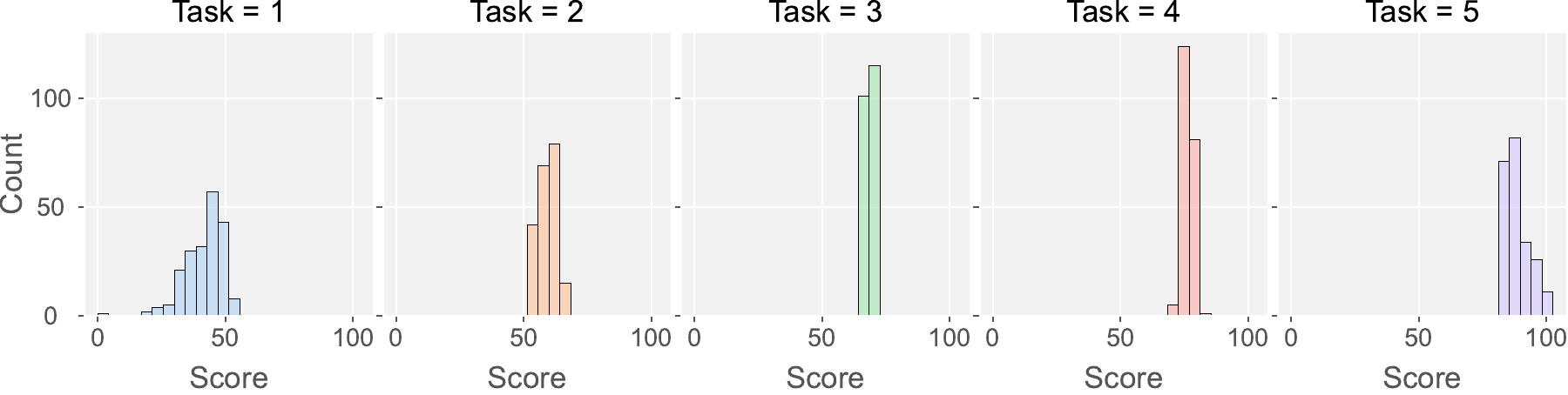}
        \put(6,21){\scriptsize Session 1}
        \put(25,21){\scriptsize Session 2}
        \put(44,21){\scriptsize Session 3}
        \put(63,21){\scriptsize Session 4}
        \put(82,21){\scriptsize Session 5}
        \end{overpic}
        
        \caption{
            Illustration of the grade-incremental setting for CAQA: Our setting is characterized by challenges of both classification and regression tasks. The interdependencies of non-stationary grades pose a new CL challenge for memorizing such dependencies.
        }
        \label{fig:cl_setting}
\end{figure}

To simulate the real-world skill variations, we propose a novel grade-incremental setting for CAQA, characterized by the challenges of both regression and classification tasks (see \cref{fig:cl_setting}). This is achieved by discretizing the continuous quality space into distinct intervals corresponding to different action grades and ensuring an equal number of samples in each session, resulting in more challenging score variations. Unlike the uniform discrete class semantic space in traditional class-incremental tasks \cite{zhang2023few}, our setting involves contextual relationships between adjacent grades, and data samples in the same grade may present quality variations of actions, posing a new challenge for lifelong learning in preserving these dependencies to mitigate catastrophic forgetting.

For MTL-AQA, FineDiving, and UNLV-Dive, we partition them into five subsets, ensuring each subset has an equal number of samples according to their label distribution. JDM-MSA, with a limited number of actions, is divided into three subsets. This split creates a challenging protocol for CAQA models to handle the variability in quality scores, allowing us to evaluate their adaptability to real-world scenarios. To address timely updates and labeling scarcity, we train models with a few samples per session, reserving the remaining samples for base session fine-tuning. During inference, all samples in their subset are evaluated for a comprehensive assessment. Additionally, we assess the effectiveness and generalization of CAQA models in learning from limited instances by varying the number of training samples per session (refer to \cref{fig:sol}).

\subsection{Evaluation Metrics} 
The Spearman's rank correlation coefficient, denoted as $\rho$ quantifies the correlation between the ground truth $\bm{y}$, and its predicted score $\hat{\bm{y}}$. Given that $\bm{p}$ and $\bm{q}$ represent the rank vectors of $\bm{y}$ and $\hat{\bm{y}}$, $\rho$ is defined as:
\begin{equation}
  \rho = \frac{ \sum_i (p_i - \bar{p}) (q_i - \bar{q}) }{\sqrt{\sum_i (p_i - \bar{p})^2 \sum_i (q_i - \bar{q})^2}},
\end{equation}
where $\bar{p}$ and $\bar{q}$ denote the average values of $\boldsymbol{p}$ and $\boldsymbol{q}$. 

To comprehensively assess the efficacy of CAQA models, we have adopted three CL metrics as outlined in \cite{wang2023comprehensive}.
We utilize the overall correlation, denoted as $\rho_{\mathrm{avg}}$, as a metric to gauge total performance. This approach differs from previous methodologies \cite{tao2020few,yang2023neural}, which tend to compute individual metrics in isolation. By aggregating samples from all preceding sessions, we aptly address the rank correlation's sensitivity to sample size variations.
Furthermore, to quantify memory stability and learning plasticity, we employ the average forgetting $\rho_{\mathrm{aft}}$ and the forward transfer $\rho_{\mathrm{fwt}}$, respectively. $\rho_{\mathrm{aft}}$ and $\rho_{\mathrm{fwt}}$ can be defined as:
\vspace{-0.1cm}
\begin{align}
\rho_{\mathrm{aft}} =  \frac{1}{T-1} & \sum_{t=1}^{T-1} \max_{i,j\in \{1,2,\cdots,T\}} \left( \rho_{i,t} - \rho_{j,t} \right), \\
\rho_{\mathrm{fwt}} = & \frac{1}{T-1} \sum_{t=2}^{T} \left( \rho_{t-1,t} - \tilde{\rho}_{t} \right),
\end{align}
where $\rho_{i,j}$ represents the correlation evaluated on the test set of the $j$-th task after incremental learning of the $i$-th task ($j \leq i$), and $\tilde{\rho}_{t}$ denotes the correlation of a randomly initialized reference model evaluated on that of the $t$-th task.

\section{Experimental Results}
Using PyTorch, all models are trained with two NVIDIA RTX 3090 GPUs. 
We adopted Adam with a learning rate and weight decay both set to $10^{-4}$. Each training session spans a maximum of 50 epochs. We set the batch size $b_1$ to 5 and the mini-batch size $b_2$ to 3 across all models. We have frozen the batch normalization layer within the backbone to counterbalance the influence of batch size. In addition, we employed two MLP layers for the MP module, and both loss weight parameters, $\lambda_{\mathrm{P}}$ and $\lambda_{\mathrm{R}}$, are set to 1. 
We present the main results here, with supplementary material offering additional details.

\myPara{Comparison with Recent Strong Baselines.} 
We compared MAGR against a series of baselines, including both memory-free methods \cite{riemer2019learning,buzzega2020dark,tao2020few,kukleva2021generalized,yang2023neural} and memory-based approaches \cite{zenke2017continual,james2017ewc,li2017learning,zhang2023slca}. Detailed results are provided in \cref{tab:cmp_sota}. 

\begin{table*}[h!]
    \centering
    \setlength{\abovecaptionskip}{-0.05cm}
    \setlength{\belowcaptionskip}{-0.01cm}
    \scriptsize
    \caption{ 
    Experimental results for CAQA models. The primary metric considered is $\rho_{\mathrm{avg}}$. 
    We opt not to incorporate the difficulty label in MTL-AQA and the dive number in FineDiving for consistency to maintain a fair evaluation protocol.
    }
    \resizebox{\linewidth}{!}{
    \begin{tabular}{lrccccccccccccc}
    \toprule
    \multirow{2}{*}[-0.5ex]{Method} & \multirow{2}{*}[-0.5ex]{Publisher} &  \multirow{2}{*}[-0.5ex]{Memory}   & \multicolumn{3}{c}{MTL-AQA} & \multicolumn{3}{c}{FineDiving} & \multicolumn{3}{c}{UNLV-Dive} & \multicolumn{3}{c}{JDM-MSA} \\ \cmidrule(r){4-6} \cmidrule(r){7-9} \cmidrule(r){10-12} \cmidrule(r){13-15}
    & & & 
    $\rho_{\mathrm{avg}}$ ($\uparrow$) & $\rho_{\mathrm{aft}}$ ($\downarrow$) & $\rho_{\mathrm{fwt}}$ ($\uparrow$) &
    $\rho_{\mathrm{avg}}$ ($\uparrow$) & $\rho_{\mathrm{aft}}$ ($\downarrow$) & $\rho_{\mathrm{fwt}}$ ($\uparrow$) &
    $\rho_{\mathrm{avg}}$ ($\uparrow$) & $\rho_{\mathrm{aft}}$ ($\downarrow$) & $\rho_{\mathrm{fwt}}$ ($\uparrow$) & $\rho_{\mathrm{avg}}$ ($\uparrow$) & $\rho_{\mathrm{aft}}$ ($\downarrow$) & $\rho_{\mathrm{fwt}}$ ($\uparrow$)  \\
    \midrule
    Joint Training & - & None & 0.9360 & - & - & 0.9075 & - & - & 0.8460 & - & - & 0.7556 & - & - \\
    \hline
    Sequential FT & - & None & 0.5458 & 0.1524 & 0.0538 & 0.7420 & 0.1322 & 0.2135 & 0.6307 & 0.2135 & \bf ~0.3595 & 0.5080 & 0.1029 & 0.5431  \\
    SI \cite{zenke2017continual} & ICML'17 & None & 0.5526 & 0.2677 & 0.0350 & 0.6863 & 0.2330 & 0.1938 & 0.1519 & 0.3822 &  ~0.0220 & 0.4804 & 0.2198 & 0.5431 \\
    EWC \cite{james2017ewc} & PNAS'17 & None & 0.2312 & 0.1553 & 0.0343 & 0.5311 & 0.3177 & 0.1776 & 0.4096 & 0.2576 & ~0.3039 & 0.3889 & 0.1690 & 0.3120 \\
    LwF \cite{li2017learning} & TPAMI'17 & None & 0.4581 & 0.1894 & 0.0490 & 0.7648 & 0.0807 & 0.2894 & 0.6081 & 0.1578 & ~0.3230 & 0.6441 & 0.1127 & 0.2423 \\
    \hline
    MER \cite{riemer2019learning} & ICLR'19 & Raw Data & 0.8720 & 0.1303 & 0.0625 & 0.8276 & 0.1446 & 0.2806 & 0.7397 & 0.1321 & ~0.0465 & 0.6689 & \bf 0.0635 & 0.3841 \\
    DER++ \cite{buzzega2020dark} & NeurIPS'20 & Raw Data & 0.8334 & 0.1775 & 0.0433 & 0.8285 & 0.1523 & 0.2851 & 0.7206 & 0.1382 & -0.1773 & 0.5364 & 0.0835 & 0.5759 \\
    TOPIC \cite{tao2020few} & CVPR'20 & Raw Data & 0.7693 & 0.1427 & 0.1391 & 0.8006 & 0.1344 & 0.2744 & 0.4085 & 0.2647 & ~0.1132 & 0.6575 & 0.2184 & 0.5492 \\
    GEM \cite{kukleva2021generalized} & ICCV'21 & Raw Data & 0.8583 & 0.0950 & 0.1429 & 0.8309 & 0.0721 & 0.2883 & 0.6538 & 0.2322 & ~0.0270 & 0.6084 & 0.0499 & 0.3566 \\
    \hline
    Feature MER & - & Feature & 0.7283 & 0.2255 & 0.0535 & 0.4914 & 0.2354 & 0.2344 & 0.5675 & 0.1322 & ~0.1558 & 0.6295 & 0.1597 & \bf 0.6446  \\
    SLCA \cite{zhang2023slca} & ICCV'23 & Feature & 0.7223 & 0.1852 & 0.1665 & 0.8130 & 0.0920 & 0.2453 & 0.5551 & 0.1085 & ~0.3200 & 0.6173 & 0.1705 & 0.4457 \\
    NC-FSCIL \cite{yang2023neural} & ICLR'23 & Feature & 0.8426 & 0.1146 & 0.0718 & 0.8087 & 0.0203 & \bf 0.3404 & 0.6458 & \bf 0.0637 & -0.1677 & 0.6571 & 0.1295 & 0.4957 \\
    \rowcolor{LightCyan} MAGR (Ours) & - & Feature & \bf 0.8979 & \bf 0.0223 & \bf 0.1914 & \bf 0.8580 & \bf 0.0167 & 0.2952 & \bf 0.7668 & 0.0827 & ~0.1227 & \bf 
 0.7166 & 0.1069 & 0.4957 \\
    \bottomrule
    \end{tabular}
    }
    \label{tab:cmp_sota}
\end{table*}

Joint training represents the upper performance bound for CL. In contrast, using sequential fine-tuning, the lower bound leads to a significant drop in performance across all datasets, such as a 41.69\% correlation decline on the MTL-AQA dataset. This verifies the challenge of catastrophic forgetting in the CAQA setting. While memory-free models make some effort, their performance pales in comparison to raw data replay and feature replay. 

The raw data replay is effective but raises privacy concerns, where MER \cite{riemer2019learning} achieves the best performance among them. Although feature replay can handle privacy concerns, it encounters challenges when the backbone undergoes fine-tuning in new sessions. This can lead to feature deviation from the current data manifold, impacting assessment performance. This issue can be confirmed when comparing the Feature MER (feature replay adaptation of MER) with that of MER, where the former performs poorly compared to the latter. NC-FSCIL \cite{yang2023neural} adopts a fixed backbone approach to avoid deviations but sacrifices adaptability to real-world complexity, resulting in lower performance than MER. Our method is specifically engineered to address feature deviation and thus achieves the best performance, underscoring its superior design and efficacy in CAQA.

We have noted that recent prompt-based approaches \cite{wang2023hierarchical} have excelled in CL on pre-trained models. However, due to the distinctive nature of AQA pre-training in terms of model architectures and datasets, the use of single-task (ViT+prompt) remains unexplored. Recently, SLCA \cite{zhang2023slca} has demonstrated superior performance to classical prompt-based approaches like DualPrompt \cite{wang2022dualprompt}. Therefore, we compare MAGR with SLCA, which performs even worse than NC-FSCIL due to its consistently lower quality of generative features. This unstable generation affects the CAQA performance in mitigating catastrophic forgetting. This emphasizes the clear advantages of our method.

\begin{wraptable}{r}{0.62\linewidth}
\vspace{-1.4\baselineskip}

    \centering
    \scriptsize 
    \caption{Ablation results on the MTL-AQA dataset.}
    \label{tab:ablation_study_mtl}
    \begin{tabular}{llll}
    \toprule
     Setting   & $\rho_{\mathrm{avg}}$ ($\uparrow$) & $\rho_{\mathrm{aft}}$ ($\downarrow$) & $\rho_{\mathrm{fwt}}$ ($\uparrow$)  \\
    \midrule
     MAGR (Ours)  & 0.8979 & 0.0223 & 0.1914 \\
     \hline
     ~~w/o MP  &  0.6949 $^{\downarrow 23\%}$ & 0.1325 $^{\uparrow 494\%}$ & 0.0814 $^{\downarrow 57\%}$ \\
     ~~w/o MP's res. link  &  0.8391 $^{\downarrow 7\%}$ & 0.0232 $^{\uparrow 4\%}$ & 0.1743 $^{\downarrow 9\%}$ \\
     \hline
     ~~w/o II-GR  & 0.8463 $^{\downarrow 6\%}$ & 0.0970 $^{\uparrow 335\%}$ & 0.1062 $^{\downarrow 45\%}$ \\
     ~~w/o J-GR  & 0.7839 $^{\downarrow 13\%}$ & 0.1053 $^{\uparrow 372\%}$ & 0.1005 $^{\downarrow 48\%}$ \\
     ~~w/o IIJ-GR  & 0.7362 $^{\downarrow 18\%}$ & 0.1232 $^{\uparrow 452\%}$ & 0.0883 $^{\downarrow 54\%}$ \\
     ~~w/o KL (MSE)  &  0.8447 $^{\downarrow 6\%}$ & 0.0265 $^{\uparrow 16\%}$ & 0.1890 $^{\downarrow 1\%}$ \\
     \hline
     ~~w/o OUS (random) & 0.8619 $^{\downarrow 4\%}$  & 0.0876 $^{\uparrow 293\%}$ &  0.1027 $^{\downarrow 46\%}$ \\
    \bottomrule
    \end{tabular}
\end{wraptable}
\myPara{Ablation Study.} 
In \cref{tab:ablation_study_mtl}, we have conducted an ablation study on the MTL-AQA dataset.
The first row represents the performance of our MAGR model. Each subsequent row delineates the performance by removing a core component from MAGR. From \cref{tab:ablation_study_mtl}, removing the MP module results in the most significant drop in performance, underlining its central role in rectifying feature deviations and ensuring that old features align with the evolving data manifold. The profound impact on results without the IIJ-GR emphasizes its essentiality in preserving regressor alignment across sessions. Separately removing Intra-Inter Graph Regularizer (II-GR) and Joint Graph Regularizer (J-GR) clarifies that both local and global regularizations are vital for realizing CAQA. Each independently enhances performance over the scenario where both are excluded, proving the distinct contributions of both local and global components. While the performance drop of removing OUS is not as pronounced upon its exclusion, the role of OUS is beyond mere metrics. It embodies the efficiency of MAGR, ensuring a smart and compact replay by selecting the most representative exemplars, thus economizing storage. 
In sum, the ablation study verifies the collective importance of each module. Each component is not just an additive piece but brings a unique aspect.

\begin{figure*}
    \centering
    \includegraphics[width=\linewidth, clip, trim=5 8 10 8]{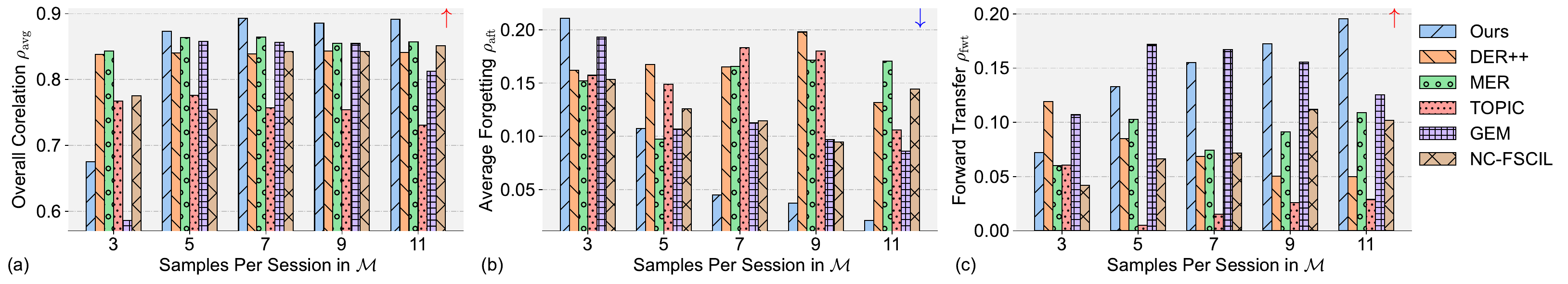}
    \caption{Memory size comparisons with replay-based methods on MTL-AQA. $\uparrow$ indicates higher values are better for the metric, whereas $\downarrow$ indicates the opposite.}{
    \phantomsubcaption\label{fig:cmp_memory_size-a}%
    \phantomsubcaption\label{fig:cmp_memory_size-b}%
    \phantomsubcaption\label{fig:cmp_memory_size-c}%
    }
    \label{fig:cmp_memory_size}
\end{figure*}

\myPara{Impact of Memory Size.}
The memory bank requires extra storage space to retain old features, constrained by limited resources. 
To trade off the performance and memory size, we varied the number of replayed samples per session in \cref{fig:cmp_memory_size}. MAGR demonstrates resilience within the range of 5 to 11 features per session, owing to the integration of OUS to sample representative features. However, when the sample number is 3, MAGR's performance degrades. This is likely because OUS may sample extreme boundary samples that fail to maintain old distributions effectively, whereas random sampling may perform better in such scenarios. We maintained 10 samples per session for a fair comparison in \cref{tab:cmp_sota}.

\begin{figure}
\centering
\includegraphics[width=\linewidth]{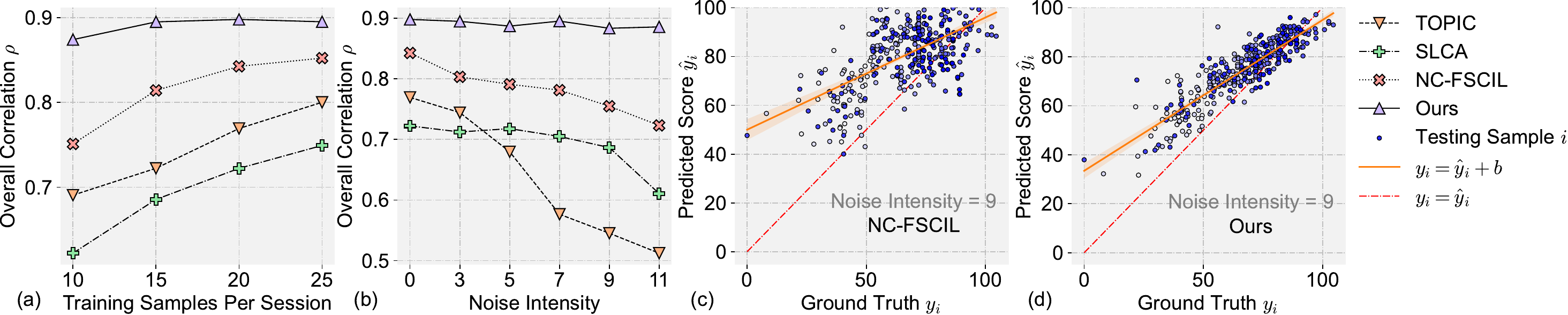}
\caption{Illustrations of label scarcity and noise robustness: (a) Training samples per session plot, (b) Noise intensity plot, (c) Correlation plot of NC-FSCIL at noise intensity 9, and (d) Correlation plot of MAGR at noise intensity 9.}{
\phantomsubcaption\label{fig:sol-a}%
\phantomsubcaption\label{fig:sol-b}%
\phantomsubcaption\label{fig:sol-c}%
\phantomsubcaption\label{fig:sol-d}%
}
\label{fig:sol}
\end{figure}

\myPara{Robustness to Label Scarcity, Noises and Severe Deviations.} 
Unlike classification, obtaining reliable quality scores for actions typically requires domain experts and specialized annotation procedures. 
This scarcity of labeled AQA data is a major challenge not addressed by most CL methods tested on plentiful classification benchmarks. 
We evaluated MAGR's performance under different label scarcity conditions (see \cref{fig:sol-a}) and examined its robustness to noise (see \cref{fig:sol-b}) on the MTL-AQA dataset. Different levels of label noise were only introduced to the training data for comparison against recent strong baselines \cite{zhang2023slca,yang2023neural,tao2020few}, showcasing MAGR's effectiveness in learning from fewer labeled examples and robustness to noise. Additionally, we have visualized the correlation plots at noise intensity 9 to intuitively compare a recent baseline \cite{yang2023neural} (see \cref{fig:sol-c}) and MAGR (see \cref{fig:sol-d}), demonstrating MAGR's resilience to varying noise intensities. 
We further quantify the feature deviations between pre-trained and fine-tuned features of Joint Training in \cref{tab:deviation} (JDM-AQA is not considered due to the different feature scales). Our correlation gains compared to the strongest baseline increase significantly with feature deviations, indicating MAGR's robustness in handling severe deviations. This is due to the graph regularization's ability to preserve the feature space structure.

\begin{table}[h]
    \centering
    \scriptsize
    \caption{Statistics of feature deviations and correlation gains.}
    \begin{tabular}{lrrr}
    \toprule
    Dataset  & FineDiving & MTL-AQA & UNLV-Dive \\ 
    \midrule
    Degree of Feature Deviations (MSE)  &  26.85 & 35.28 & 51.75 \\ 
    Overall Correlation Gains (\%) &  5.66 & 6.56 & 15.64 \\ \bottomrule
    \end{tabular}
    
    \label{tab:deviation}
\end{table}

\begin{figure}[!h]
    \centering
    \includegraphics[width=\linewidth]{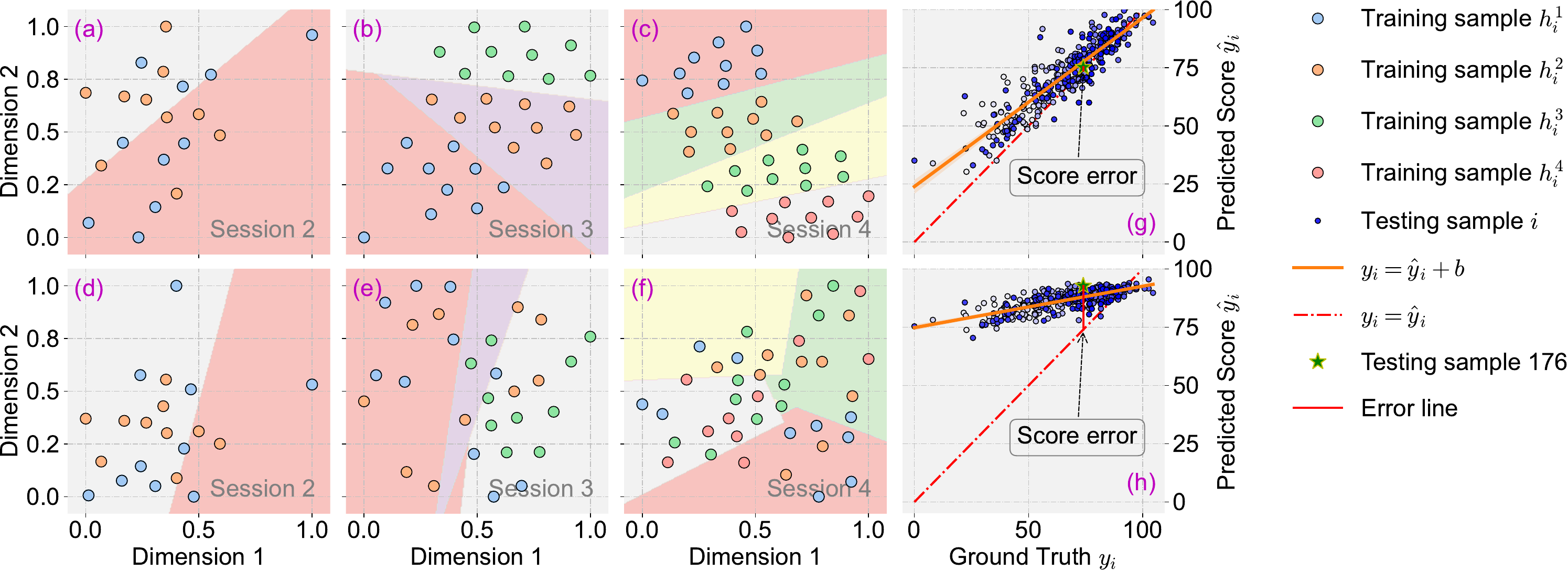}
    \caption{Visualizations of feature distribution (a-f) and score correlation (g-h): MAGR (top) and Feature MER (bottom).
    The explicit division of different sessions validates the effectiveness of MAGR in mitigating catastrophic forgetting.
    }{
        \phantomsubcaption\label{fig:tsne-a}%
        \phantomsubcaption\label{fig:tsne-b}%
        \phantomsubcaption\label{fig:tsne-c}%
        \phantomsubcaption\label{fig:tsne-d}%
        \phantomsubcaption\label{fig:tsne-e}%
        \phantomsubcaption\label{fig:tsne-f}%
        \phantomsubcaption\label{fig:tsne-g}%
        \phantomsubcaption\label{fig:tsne-h}%
    }
    \label{fig:tsne}
\end{figure}

\begin{figure}
    \centering
    \begin{overpic}[width=0.16\linewidth,height=0.16\linewidth]{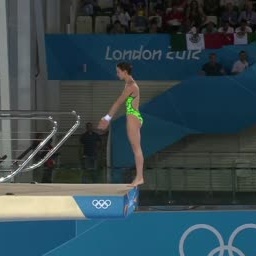}
        \put(5,88){\tiny\color{yellow}\#176}
        \put(62,88){\tiny\color{yellow}01-st}
        
        \put(0,0){\tikz\fill[white,fill opacity=0.5](0,0)rectangle(0.4,0.4);}
        \put(2,8){\tiny \color{black}(a)}
    \end{overpic}
    \begin{overpic}[width=0.16\linewidth,height=0.16\linewidth]{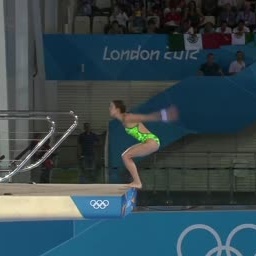}
        \put(5,88){\tiny\color{yellow}\#176}
        \put(62,88){\tiny\color{yellow}31-st}
        
        \put(0,0){\tikz\fill[white,fill opacity=0.5](0,0)rectangle(0.4,0.4);}
        \put(2,8){\tiny \color{black}(b)}
    \end{overpic}
    \begin{overpic}[width=0.16\linewidth,height=0.16\linewidth]{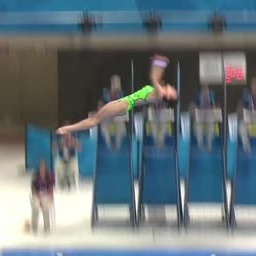}
        \put(5,88){\tiny\color{yellow}\#176}
        \put(62,88){\tiny\color{yellow}51-st}
        
        \put(0,0){\tikz\fill[white,fill opacity=0.5](0,0)rectangle(0.4,0.4);}
        \put(2,8){\tiny \color{black}(c)}
    \end{overpic}
    \begin{overpic}[width=0.16\linewidth,height=0.16\linewidth]{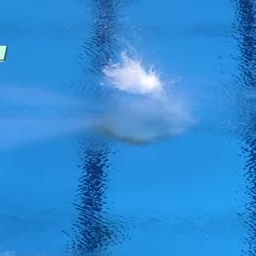}
        \put(5,88){\tiny\color{yellow}\#176}
        \put(62,88){\tiny\color{yellow}71-st}
        
        \put(0,0){\tikz\fill[white,fill opacity=0.5](0,0)rectangle(0.4,0.4);}
        \put(2,8){\tiny \color{black}(d)}
    \end{overpic}
    \begin{overpic}[width=0.16\linewidth,height=0.16\linewidth]{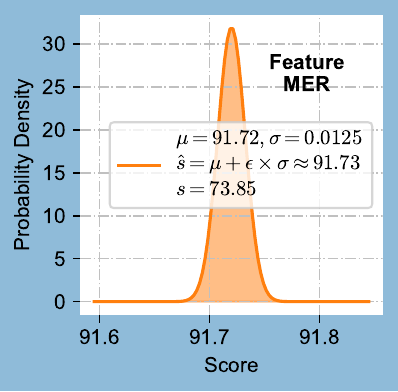}
        \put(2,8){\tiny \color{black}(e)}
    \end{overpic}
    \begin{overpic}[width=0.16\linewidth,height=0.16\linewidth]{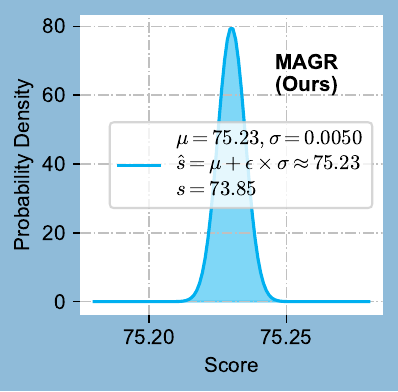}
        \put(2,8){\tiny \color{black}(f)}
    \end{overpic}
    \caption{
    Qualitative assessment comparison of sample \#176 (a-d) using Feature MER (e) and MAGR (f) on the MTL-AQA dataset. We adopted the reparameterization trick \cite{zhou2023hierarchical} to obtain the predicted score, where $\mu$ and $\sigma$ represent the mean and the standard deviation (please see our supplementary material for details).
    }{
    \phantomsubcaption\label{fig:example-a}%
    \phantomsubcaption\label{fig:example-b}%
    \phantomsubcaption\label{fig:example-c}%
    \phantomsubcaption\label{fig:example-d}%
    \phantomsubcaption\label{fig:example-e}%
    \phantomsubcaption\label{fig:example-f}%
    }
    \label{fig:example}
\end{figure}
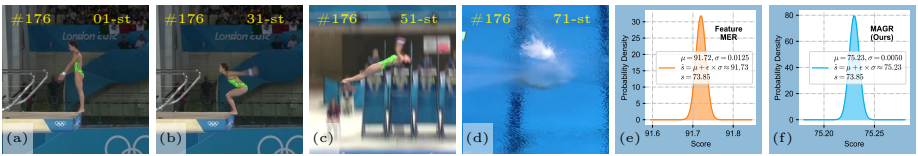

\myPara{Visualization of Mitigating Catastrophic Forgetting.}
We employed t-SNE \cite{van2008visualizing} to project the features into the 2D space on the MTL-AQA dataset. \cref{fig:tsne-c} highlights the efficacy of MAGR in organizing samples across various sessions coherently, while Feature MER struggles with feature deviation (see \cref{fig:tsne-f}). We further showcase correlation plots at the end of the last session in the last column. From \cref{fig:tsne-g} and \cref{fig:tsne-h}, it is evident to demonstrate our superior correlation between the predicted scores and the actual ground truth, evaluating the effectiveness of maintaining feature alignment across sessions. For example, the ground truth score of sample \#176 (see \cref{fig:example}) is 73.85. Our method's prediction remains within a margin of 2 points from the ground truth score, indicating a closer alignment compared to Feature MER.

\section{Conclusions and Future Work}
This work proposes the novel task of CAQA to accommodate real-world complexities. To mitigate catastrophic forgetting while prioritizing user privacy, we propose MAGR as a solution to address the misalignment issue due to backbone updates. By integrating MP and IIJ-GR, MAGR iteratively refines deviated old features and regulates the feature space across incremental sessions. Experiments on three AQA datasets show the superiority of MAGR compared to recent strong baselines. 
We believe this to enhance AQA systems in real-world applications, offering improved capabilities to serve our human beings.
Future research should explore more robust designs capable of handling complex data. This may involve optimizing the number of layers and investigating advanced network architectures like ViT. In this way, incorporating prompt-based techniques could be considered to ensure parameter-efficient tuning and enhance adaptability.
\setcounter{page}{1}
\setcounter{table}{0}
\setcounter{figure}{0}

\renewcommand{\theequation}{\textcolor{gray}{S}\arabic{equation}}
\renewcommand{\thetable}{\textcolor{gray}{S}\arabic{table}}
\renewcommand{\thefigure}{\textcolor{gray}{S}\arabic{figure}}

\appendix

\section{Details of the AQA Network}

\begin{figure}[h]
    \centering
    \includegraphics[width=0.6\linewidth,clip,trim=140 130 140 120]{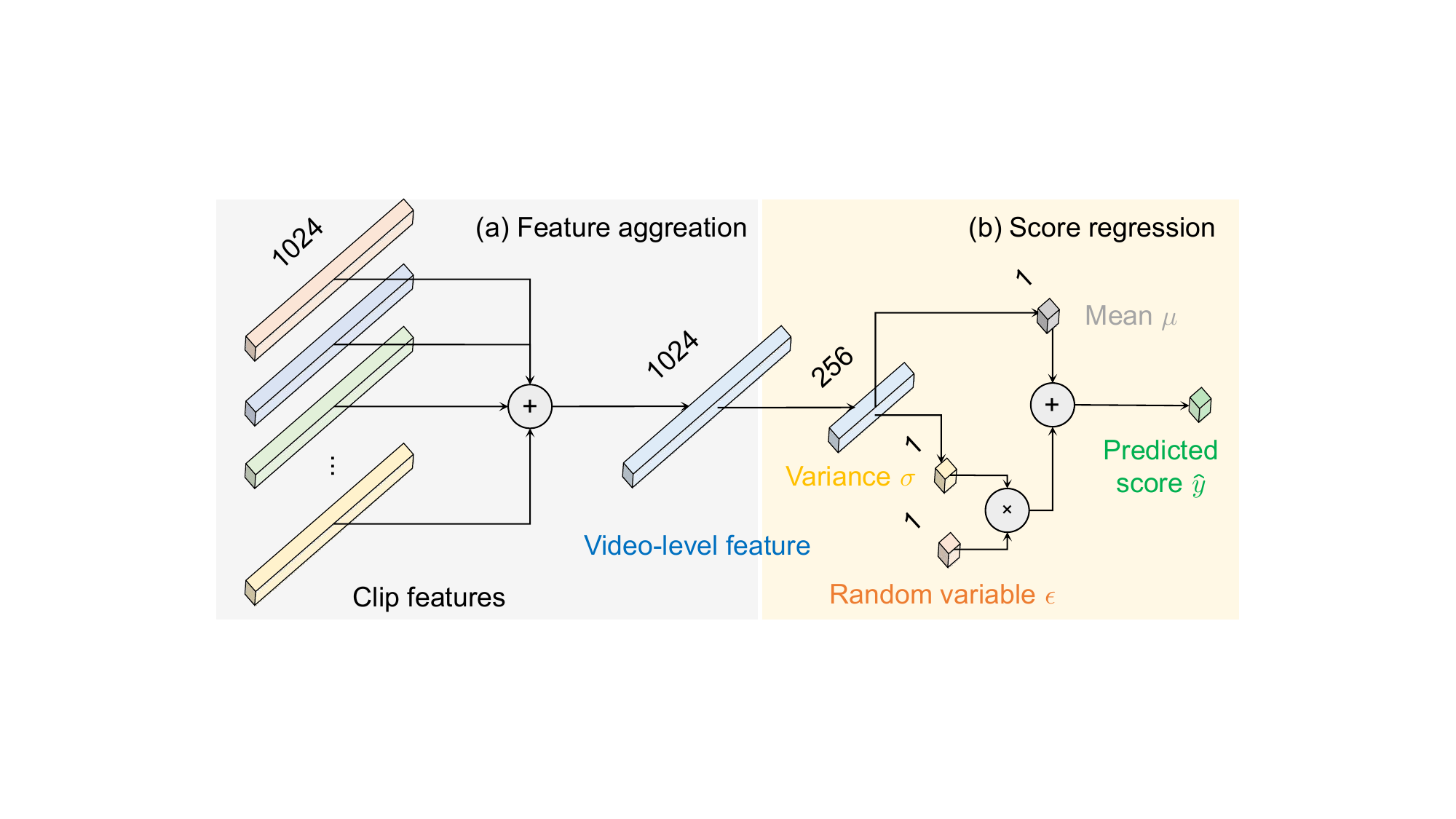}
    \caption{The network architecture for score regression: (a) depicts feature aggregation and (b) delineates score regression.}{
    \phantomsubcaption\label{fig:detail_arch-a}%
    \phantomsubcaption\label{fig:detail_arch-b}%
    }
    \label{fig:detail_arch}
\end{figure}

\cref{fig:detail_arch} illustrates the network architecture of our AQA framework, employing the {\tt I3D+MLP} paradigm. Following the previous work \cite{zhou2023hierarchical}, we introduce the re-parameterization technique \cite{kingma2019introduction} to ensure robust score regression. 

\myPara{Feature Aggregation.} 
To address computational challenges in Action Quality Assessment (AQA), methods like \cite{yu2021group,zhou2023hierarchical,bai2022action,pan2019action,parmar2019action} often opt to divide videos into clips. This process involves uniformly dividing the entire video sequence $\mathbf{x}$ into 10 clips $\mathbf{c}_1, \mathbf{c}_2, \cdots, \mathbf{c}_{10}$. Each of these clips is then fed into the I3D backbone to extract clip features. The division helps manage the computational intensity associated with processing large video datasets, enabling more efficient computation and improved memory utilization. 

These clip features are aggregated using the widely used average pooling method to obtain the whole video-level representation $\bm{h}$. Thus, the aforementioned process can be represented as:
\begin{equation}
    \bm{h} = \text{\tt AvgPool}( \text{\tt i3d} (\mathbf{c}_i), \text{\tt i3d} (\mathbf{c}_2), \cdots, \text{\tt i3d} (\mathbf{c}_{10})),
\end{equation}
where $\text{\tt AvgPool} (\cdot)$ denotes the average pooling function.
While the simple process is effective, it has limitations, such as its coarse-grained temporal approach, which may not capture fine-grained action quality \cite{zhou2023hierarchical,gedamu2023fine}. 
It is noted that our main focus is on the integration of CL and AQA, and addressing these limitations is out of our work.

\myPara{Score Regression.} 
The video-level representation is then utilized in MLPs for the final score regression. The detailed illustration (see \cref{fig:detail_arch-b}) provides clarity on the sequential steps involved in the score regression process. 

We employ a probabilistic layer to transform the video-level feature $\bm{h}$ into a random score variable $y$. The encoded score variable $y$ follows a Gaussian distribution defined by:
\begin{equation}
p({y}; \bm{h})=\frac{1}{\sqrt{2 \pi \sigma^2(\bm{h})} } \exp \left(-\frac{(y-\mu(\bm{h}))^{2}}{2 \sigma^{2}(\bm{h})}\right),
\end{equation}
where $\mu$ and $\sigma^2$ are the mean and variance parameters with respect to the feature representation. These parameters quantify the quality and uncertainty of the action score, respectively. To sample from the distribution, we apply the re-parameterization trick \cite{kingma2019introduction}, which involves sampling from another random variable $\epsilon$ following the standard normal distribution $\mathcal{N}(0,1)$. In this way, the predicted score $\hat{y}$ can be calculated as:
\begin{equation}
\hat{y}=\mu(\bm{h}) + \epsilon \cdot \sigma(\bm{h}),
\end{equation} 
where $\sigma(\cdot)$ represents the standard deviation. This ensures that the score distribution sampling process is differentiable, allowing feasible training of our score regression network.

\section{Additional Experimental Details}
This section provides supplementary information on the datasets and implementation details used in our experiments.

\subsection{Deatils of Datasets}
\myPara{MTL-AQA} \cite{parmar2019action} serves as a robust resource for AQA research, offering a comprehensive collection of 16 distinct diving events. With a total of 1412 samples, it encompasses a wide range of scenarios, featuring both male and female athletes participating in single and double diving competitions across 3 m springboard and 10 m platform categories. Notably, MTL-AQA provides detailed annotations for action categories and commentary alongside AQA scores, enhancing the dataset's utility for comprehensive analysis. 
A total of 1059 samples were allocated for training purposes, while 353 samples were reserved for testing.

\myPara{FineDiving} \cite{xu2022finediving} is a recently introduced large-scale fine-grained diving dataset comprising 3000 diving samples extracted from various prestigious events including the Olympics, World Cup, World Championships, and European Aquatics Championships. This dataset covers 52 action types, 29 sub-action types, and 23 difficulty degree types, providing rich annotations for detailed analysis. Additionally, FineDiving includes fine-grained annotations such as action type, sub-action type, coarse-grained and fine-grained temporal boundaries, and action scores in addition to AQA scores. A total of 75\% of the samples were allocated for training purposes, while the remaining 25\% were reserved for testing.

\myPara{UNLV-Dive} \cite{parmar2017learning} comprises 300 training videos and 70 testing videos, selected from the (semi-) final of the 10-meter platform diving event in the 2012 London Olympics. The dataset's final scores range from 21.6 to 102.6, with execution scores falling within the [0, 30] range. 

\myPara{JDM-MSA} \cite{zhou2023video} comprises 14 types of actions categorized into three groups: time-related actions, position-related actions, and uncertain actions. Time-related actions are evaluated based on completion duration, and position-related actions on body part movement magnitude and position, while uncertain actions require subjective judgments of difficulty. For this study, we focus on six challenging actions, prioritizing those requiring subjective assessment. Data collection efficiency was enhanced using an iPad, an iPhone, and two USB cameras. To ensure consistency, all samples were normalized to the same resolution and frame count.

\subsection{Implementation Details}
In the detailed architecture presented in \cref{fig:detail_arch}, the video-level representation undergoes an initial encoding process, resulting in a 1024-dimensional vector. This vector then traverses a series of Multi-Layer Perceptrons (MLPs). The first MLP layer, with a dimensionality of 256, serves as the initial transformation. Following this, two additional MLP layers process the vector further, ultimately mapping it to mean and variance scores. These mean and variance parameters play a crucial role in shaping the final prediction.
The MP module in MAGR follows a similar structure with a 2-layer MLP. The input dimension is set to 1024, the hidden size is 256, and the output size remains at 256. This design choice aims to capture the intricate relationships within the data manifold, enhancing the ability to align features with evolving data distributions across sessions.

\section{Additional Experiments}
This section delves into additional computational analysis, online performance, the inclusion of difficulty degree, additional ablation study, the impact of memory size, and offers visualizations to further verify the effectiveness of our method.

\subsection{Model Size and Computational Time Comparison} 
We conducted measurements on both model size (backbone + regressor) and offline training time (with the same hyper-parameter setting using 2 Nvidia RTX 3090 GPUs in distributed parallel computing on the same machine).
The results are reported in \cref{tab:computation}.
It can be seen that our method employs comparable model size and computational time while achieving much better performance than the most recent strong baselines. Specifically, the inclusion of a manifold projector slightly increases our model size compared to these baselines, while our training time remains competitive.

\begin{table}[h]
    \small
    \centering
    \caption{Computational performance on the MTL-AQA dataset.}
    \begin{tabular}{lccccc}
    \toprule
    \multirow{2}{*}{Method}    &  \multirow{2}{*}{\makecell{Param. \\ (M)}} & \multirow{2}{*}{\makecell{Training\\Time (h)}} &  \multicolumn{3}{c}{Offline Performance}  \\  \cline{4-6}
    &&&$\rho_{\mathrm{avg}}$ ($\uparrow$) & $\rho_{\mathrm{aft}}$ ($\downarrow$) & $\rho_{\mathrm{fwt}}$ ($\uparrow$) \\ \hline
    SLCA \cite{zhang2023slca} & 13.62 & 2.27 &0.7223 &0.1852& 0.1665 \\
    NC-FSCIL \cite{yang2023neural} & 12.62 & 2.33 & 0.8426 &0.1146& 0.0718\\
    Feature MER & 12.62 & 2.22 & 0.7283 & 0.2255 & 0.0535 \\ \hline
    MAGR (Ours)  & 12.63 & 2.23 & 0.8979 & 0.0223 & 0.1914\\ \bottomrule
    \end{tabular}
    \label{tab:computation}
\end{table}

\subsection{Real-Time Performance} 
In our online continual learning setting, we evaluate the performance of models in real-time scenarios where data arrives sequentially and models need to learn and adapt continuously without revisiting previous data. Each method is trained and updated in an online manner, processing one data point or a small batch at a time (training 1 epoch for all models), simulating real-world conditions where retraining on the entire dataset is infeasible. 
The results in \cref{tab:offline} highlight the superior online CL performance of MAGR, which achieves the highest average correlation ($\rho_{\mathrm{avg}}$) across all datasets: MTL-AQA (0.5196), FineDiving (0.4641), UNLV-Dive (0.4202), and JDM-MSA (0.2029). These results demonstrate that MAGR's graph regularization effectively preserves the feature space structure, making it a strong contender for real-time action assessment.
\begin{table}[]
    \centering
    \scriptsize
    \caption{Online continual learning ($\rho_{\mathrm{avg}}$ is the main metric).}
        \resizebox{1\linewidth}{!}{
        \begin{tabular}{lcccccccccccc}
        \toprule
        \multirow{2}{*}[-0.5ex]{Method}  & \multicolumn{3}{c}{MTL-AQA} & \multicolumn{3}{c}{FineDiving} & \multicolumn{3}{c}{UNLV-Dive} & \multicolumn{3}{c}{JDM-MSA} \\ \cmidrule(r){2-4} \cmidrule(r){5-7} \cmidrule(r){8-10} \cmidrule(r){11-13}   
        &
    $\rho_{\mathrm{avg}}$ ($\uparrow$) & $\rho_{\mathrm{aft}}$ ($\downarrow$) & $\rho_{\mathrm{fwt}}$ ($\uparrow$) &
    $\rho_{\mathrm{avg}}$ ($\uparrow$) & $\rho_{\mathrm{aft}}$ ($\downarrow$) & $\rho_{\mathrm{fwt}}$ ($\uparrow$) &
    $\rho_{\mathrm{avg}}$ ($\uparrow$) & $\rho_{\mathrm{aft}}$ ($\downarrow$) & $\rho_{\mathrm{fwt}}$ ($\uparrow$) & $\rho_{\mathrm{avg}}$ ($\uparrow$) & $\rho_{\mathrm{aft}}$ ($\downarrow$) & $\rho_{\mathrm{fwt}}$ ($\uparrow$)  \\
            \midrule
            SLCA \cite{zhang2023slca}    & 0.4880 & 0.0430 & -0.0282 & 0.3935 & 0.3360 & 0.2346 & 0.3119 & 0.1641 & -0.3082 & 0.1726 & 0.0589 & 0.0382\\
            NC-FSCIL \cite{yang2023neural}    & 0.4971 & 0.0291 & -0.0463 & 0.3810 & 0.0079 & \textbf{0.2518} & 0.3136 & \textbf{0.1282} & -0.4892 & 0.1540 & \textbf{0.0355} & 0.0378 \\
            Feature MER & 0.3571 & 0.1444 & \textbf{-0.0213} & 0.1935 & 0.0998 & 0.1559 & 0.1308 & 0.2126 & -0.4571 & 0.1699 & 0.0356 & 0.0382 \\ \hline
            MAGR (Ours) & \textbf{0.5196} & \textbf{0.0269} &  -0.0337 & \textbf{0.4641} & \textbf{0.0062} & 0.2020 & \textbf{0.4202} & 0.1947 & \textbf{-0.0499} & \textbf{0.2029} & 0.0356 & \textbf{0.0449} \\ 
            \bottomrule
        \end{tabular}
        }
    
    \label{tab:offline}
\end{table}

\subsection{Comparison with the Inclusion of Difficulty Degree}
Previous research \cite{zhou2023hierarchical,yu2021group,bai2022action} has indicated that integrating difficulty degree labels can significantly enhance the performance of AQA models, particularly on the MTL-AQA dataset. In alignment with this observation, we conducted experiments incorporating difficulty degree labels in our evaluation, as depicted in \cref{tab:cmp_sota_w_dd}. The setting is the same as the previous work \cite{yu2021group}. The results reconfirm the beneficial impact of leveraging difficulty degree information on AQA performance. Notably, MAGR maintains its superiority over recent strong baselines, reaffirming its effectiveness in addressing AQA challenges.

\begin{table*}[h!]
    \centering
    \small
    \caption{
    Results on the MTL-AQA dataset with the difficulty degree.
    } 
    \begin{tabular}{lrcccccccccc}
    \toprule
    \multirow{2}{*}[-0.5ex]{Method} & \multirow{2}{*}[-0.5ex]{Publisher} &  \multirow{2}{*}[-0.5ex]{Memory}   & \multicolumn{3}{c}{MTL-AQA} \\ \cmidrule(r){4-6}
    & & & $\rho_{\mathrm{avg}}$ ($\uparrow$) & $\rho_{\mathrm{aft}}$ ($\downarrow$) & $\rho_{\mathrm{fwt}}$ ($\uparrow$) \\
    \midrule
    Joint Training & - & None & 0.9587 & - & -  \\
    \hline
    Sequential FT & - & None & 0.8684 & 0.1418 & 0.2282   \\
    EWC \cite{james2017ewc} & PNAS'17 & None & 0.8625 & 0.1267 & 0.1776  \\
    LwF \cite{li2017learning} & TPAMI'17 & None & 0.7852 & 0.1501 & 0.0912  \\
    \hline
    DER++ \cite{buzzega2020dark} & NeurIPS'20 & Raw Data & 0.9037 & 0.1230 & 0.3122 \\
    TOPIC \cite{tao2020few} & CVPR'20 & Raw Data & 0.8782 & 0.1394 & 0.2304 \\
    \hline
    SLCA \cite{zhang2023slca} & ICCV'23 & Feature & 0.6885 & 0.2029 & 0.0958 \\
    NC-FSCIL \cite{yang2023neural} & ICLR'23 & Feature & 0.9034 & 0.0878 & 0.1456 \\
    \rowcolor{LightCyan} MAGR (Ours) & - & Feature & \bf 0.9237 & \bf 0.0615 & \bf 0.1944  \\
    \bottomrule
    \end{tabular}
    \label{tab:cmp_sota_w_dd}
\end{table*}

\subsection{Ablation Study}
The ablation results in \cref{tab:ablation_study_aqa7,tab:ablation_study_jdm} on both the UNLV-Dive and JDM-MSA datasets demonstrate the vital role each component our MAGR model. 

\myPara{Ablation Study on UNLV-Dive.}
From the results in \cref{tab:ablation_study_aqa7}, it is evident that each component of our MAGR model plays a crucial role in its performance. Omitting the MP module leads to a 21\% decline in $\rho_{\mathrm{avg}}$ and a marked 69\% increase in $\rho_{\mathrm{aft}}$, emphasizing its significance in managing feature deviation. The graph regularizers, both local (II-GR) and global (J-GR), when removed individually or together, induce notable reductions in $\rho_{\mathrm{avg}}$ and substantial increments in $\rho_{\mathrm{aft}}$, emphasizing their essential role in regularizing the feature space. Lastly, without the OUS strategy, we observe a decrease in all performance metrics, underlining its importance in model robustness across sessions. The collective results underscore the intertwined significance of all the components in achieving the peak performance of MAGR on the UNLV-Dive dataset.

\begin{table}[!h]
    \centering
    \small
    \caption{Ablation results on the UNLV-Dive dataset.}
    \label{tab:ablation_study_aqa7}
    \begin{tabular}{llll}
    \toprule
     Setting   & $\rho_{\mathrm{avg}}$ ($\uparrow$) & $\rho_{\mathrm{aft}}$ ($\downarrow$) & $\rho_{\mathrm{fwt}}$ ($\uparrow$)  \\
    \midrule
     MAGR (Ours)  & 0.7668 & 0.0827 & 0.1227 \\
     \hline
     ~~w/o MP  &  0.6026 $^{\downarrow 21\%}$ & 0.1396 $^{\uparrow 69\%}$ & 0.1075 $^{\downarrow 12\%}$ \\
     \hline
     ~~w/o II-GR  & 0.7189 $^{\downarrow 6\%}$ & 0.1726 $^{\uparrow 109\%}$ & 0.1226 $^{\downarrow 0\%}$ \\
     ~~w/o J-GR  & 0.6549 $^{\downarrow 15\%}$ & 0.1267 $^{\uparrow 53\%}$ & 0.1466 $^{\uparrow 20\%}$ \\
     ~~w/o IIJ-GR  & 0.6261 $^{\downarrow 18\%}$ & 0.2102 $^{\uparrow 154\%}$ & 0.1442 $^{\uparrow 18\%}$ \\
     \hline
     ~~w/o OUS  & 0.7356 $^{\downarrow 4\%}$  & 0.0599 $^{\downarrow 28\%}$ &  0.0867 $^{\downarrow 29\%}$ \\
    \bottomrule
    \end{tabular}
\end{table}

\myPara{Ablation Study on JDM-MSA.}
The ablation study on the JDM-MSA dataset provides insight into the importance of the components in our MAGR model. The results demonstrate that each component significantly contributes to the model's performance. Omitting the MP module resulted in a 20\% decrease in $\rho_{\mathrm{avg}}$ and a 10\% decrease in $\rho_{\mathrm{aft}}$, highlighting its role in addressing feature deviation. Similarly, removing the graph regularizers, including II-GR, J-GR, and IIJ-GR, led to notable reductions in $\rho_{\mathrm{avg}}$, emphasizing their essential role in regularizing the feature space. The proposed OUS strategy proved to be crucial for maintaining model performance, with its removal resulting in performance declines across all metrics. These findings emphasize the critical nature of each component for achieving optimal AQA performance on the JDM-MSA dataset.

\begin{table}[!h]
\centering
\small
\caption{Ablation results on the JDM-MSA dataset.}
\label{tab:ablation_study_jdm}
\begin{tabular}{llll}
\toprule
Setting & $\rho_{\mathrm{avg}}$ ($\uparrow$) & $\rho_{\mathrm{aft}}$ ($\downarrow$) & $\rho_{\mathrm{fwt}}$ ($\uparrow$) \\
\midrule
MAGR (Ours) & 0.7166 & 0.1069 & 0.4957 \\
\hline
~~w/o MP & 0.5725 $^{\downarrow 20\%}$ & 0.1185 $^{\downarrow 10\%}$ & 0.4942 $^{\downarrow 0\%}$ \\
\hline
~~w/o II-GR & 0.6755 $^{\downarrow 6\%}$ & 0.1962 $^{\uparrow 83\%}$ & 0.4956 $^{\downarrow 0\%}$ \\
~~w/o J-GR & 0.6066 $^{\downarrow 15\%}$ & 0.0933 $^{\downarrow 13\%}$ & 0.3953 $^{\downarrow 20\%}$ \\
~~w/o IIJ-GR & 0.5792 $^{\downarrow 19\%}$ & 0.1055 $^{\downarrow 10\%}$ & 0.4085 $^{\downarrow 18\%}$ \\
\hline
~~w/o OUS & 0.6880 $^{\downarrow 4\%}$ & 0.1280 $^{\downarrow 16\%}$ & 0.4945 $^{\downarrow 0\%}$ \\
\bottomrule
\end{tabular}
\end{table}

\begin{figure*}[!h]
    \centering
    \includegraphics[width=\linewidth, clip, trim=5 8 10 10]{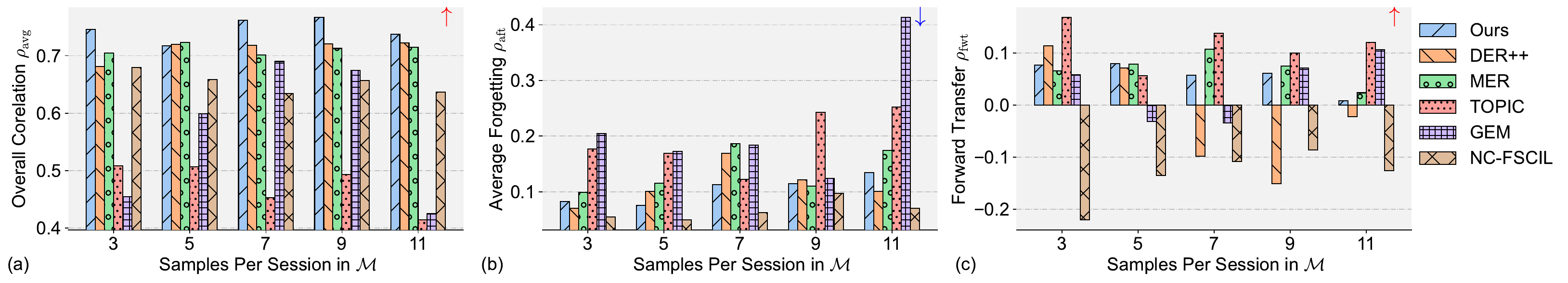}
    \caption{Memory size comparisons with replay-based methods on UNLV-Dive.}
    \label{fig:cmp_memory_size-unlv}
\end{figure*}

\begin{figure*}[!h]
    \centering
    \includegraphics[width=\linewidth, clip, trim=5 8 10 10]{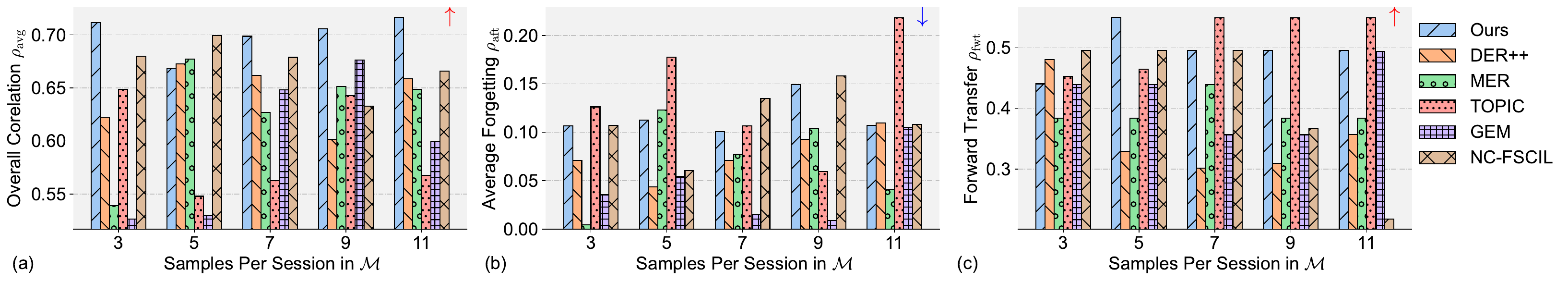}
    \caption{Memory size comparisons with replay-based methods on JDM-MSA.}
    \label{fig:cmp_memory_size-jdm}
\end{figure*}

\subsection{Impact on the Memory Size}
The performance comparison of various replay methods on the UNLV-Dive and JDM-MSA datasets for different memory sizes is presented in \cref{fig:cmp_memory_size-unlv,fig:cmp_memory_size-jdm}, respectively. On the UNLV-Dive dataset, our method consistently outperforms other approaches, achieving the highest overall correlation ($\rho_{\mathrm{avg}}$). When memory size is small, all methods experience performance degradation. However, as the memory size increases, our method exhibits superior resilience and maintains strong performance, while the performance of other methods tends to saturate or even decline. On the JDM-MSA dataset, similar trends are observed, with our method achieving the highest $\rho_{\mathrm{avg}}$ and demonstrating greater stability as memory size increases. These results highlight the effectiveness and robustness of our approach across varying memory sizes and datasets.

\subsection{Visualization of Mitigating Catastrophic Forgetting}
\begin{figure}[h]
    \centering
    \begin{overpic}[width=0.9\linewidth]{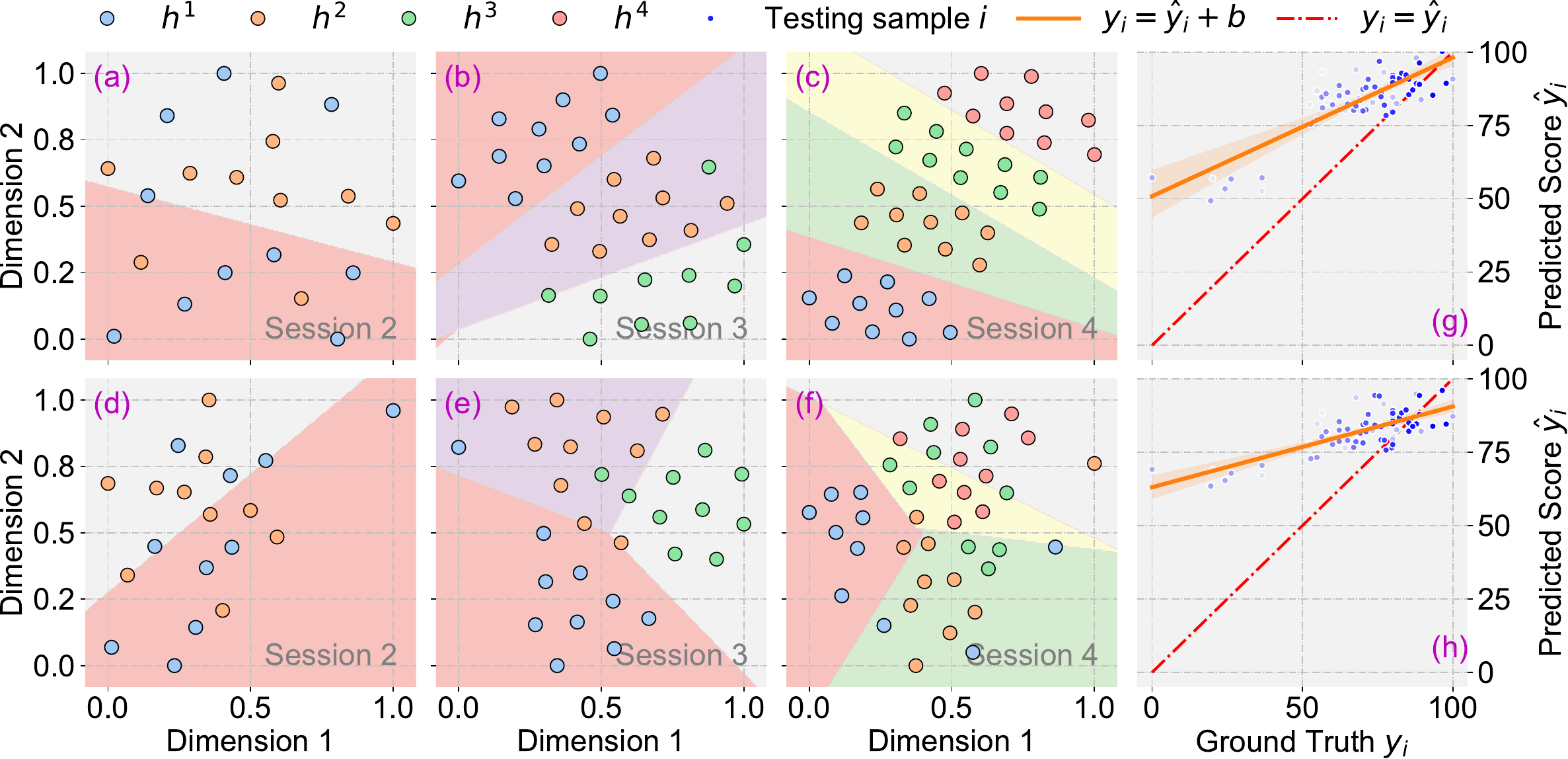}
    \end{overpic}
    \caption{Visualizations of feature distribution (a-f) and score correlation (g-h) on the UNLV-Dive dataset: MAGR (top) and Feature MER (bottom).}{
    \phantomsubcaption\label{fig:catastrophic_forgetting_visualization-a}%
    \phantomsubcaption\label{fig:catastrophic_forgetting_visualization-b}%
    \phantomsubcaption\label{fig:catastrophic_forgetting_visualization-c}%
    \phantomsubcaption\label{fig:catastrophic_forgetting_visualization-d}%
    \phantomsubcaption\label{fig:catastrophic_forgetting_visualization-e}%
    \phantomsubcaption\label{fig:catastrophic_forgetting_visualization-f}%
    \phantomsubcaption\label{fig:catastrophic_forgetting_visualization-g}%
    \phantomsubcaption\label{fig:catastrophic_forgetting_visualization-h}%
    }
    \label{fig:catastrophic_forgetting_visualization}
\end{figure}

The visualization presented in \cref{fig:catastrophic_forgetting_visualization} offers valuable insights into the effectiveness of MAGR in mitigating catastrophic forgetting on the UNLV-Dive dataset. The TSNE plots (\cref{fig:catastrophic_forgetting_visualization-a,fig:catastrophic_forgetting_visualization-b,fig:catastrophic_forgetting_visualization-c,fig:catastrophic_forgetting_visualization-d,fig:catastrophic_forgetting_visualization-e,fig:catastrophic_forgetting_visualization-f}) showcase the distribution of feature representations learned by MAGR (top) and feature MER (bottom) across different sessions. It is evident that MAGR's MP and IIJ-GR contribute to maintaining the consistency and continuity of feature distributions over sequential sessions, thereby alleviating the adverse effects of catastrophic forgetting. Notably, the correlation plots (\cref{fig:catastrophic_forgetting_visualization-g,fig:catastrophic_forgetting_visualization-h}) further emphasize MAGR's superior performance in preserving the correlation between predicted quality scores and ground truth labels, even amidst evolving feature distributions. This highlights MAGR's robustness and efficacy in addressing the challenges posed by non-stationary data distributions and underscores its potential to enhance CL and AQA research.

\subsection{Visualization of Performance Changes}

\begin{figure}[!h]
    \centering
    \includegraphics[width=0.8\linewidth, clip, trim=5 8 5 8]{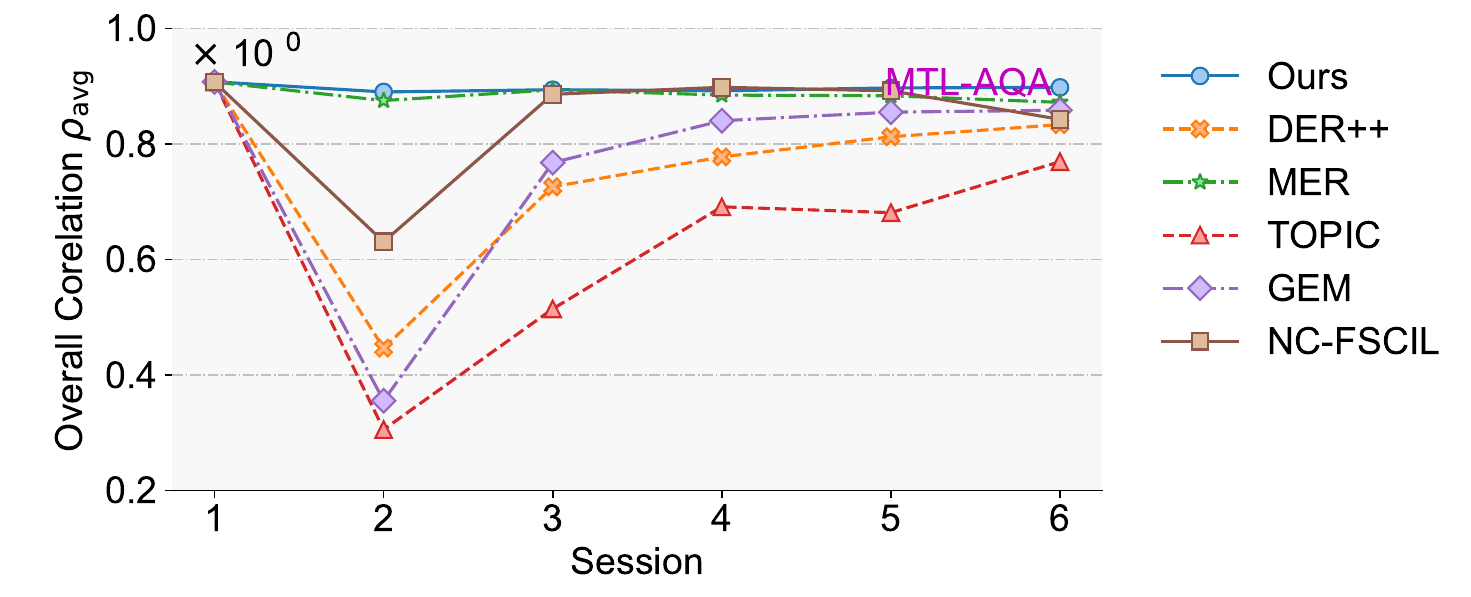}
    \caption{Performance comparison across different sessions.}
    \label{fig:cmp_session-mtl}
\end{figure}

In \cref{fig:cmp_session-mtl}, the performance changes ($\rho_{\mathrm{avg}}$) of several recent strong baselines across different sessions on the MTL-AQA dataset are illustrated. Initially, all methods exhibit comparable performance. However, as sessions progress, our method consistently maintains a stable level, while other methods experience substantial fluctuations. This underscores the effectiveness of our approach in balancing learning plasticity and memory stability over sequential sessions.

\section{Discussions}
\myPara{Larger Dataset Validation.}
We evaluated our method, MAGR, across four AQA datasets of varying scales, consistently achieving leading performance. Given the typically scarce training samples for AQA tasks, FineDiving is one of the largest available datasets. We plan to update our results with larger AQA datasets as they become available.

\myPara{Overfitting with Smaller Datasets.}
We acknowledge the potential risk of overfitting with smaller datasets or fewer feature points, as noted. To address this, we conducted experiments with reduced training samples, as shown in Fig. 8(a) of our main paper. When reducing the number of training samples per session, the performance of several recent strong baselines, especially those with non-graph feature replay methods like SLCA \cite{zhang2023slca} and NC-FSCIL \cite{yang2023neural}, decreases significantly. In contrast, our method maintains relatively stable performance, resulting in a significant performance lead. This stability is attributed to the use of graph construction, which captures and retains relationships between feature points, thereby reducing the risk of overfitting.

\myPara{Handling Difficult Cases of Actions and Qualities.}
We also analyzed the specific errors related to actions and qualities in our main paper. \cref{fig:cmp_session-mtl} evaluates the performance of specific actions learned in respective sessions. Our performance remains consistently high across different actions, whereas other strong baselines experience performance degradation in specific actions, particularly in session 2. Additionally, Fig. 9(g) and 9(h) evaluate the performance of specific qualities of each action. Compared to Feature MER in Fig. 9(h), our method, shown in Fig. 9(g), aligns much better with the ground truth, indicating fewer specific errors. While our method shows slight deviations within the low-score area due to limited training samples, the high-score area, which is densely sampled, is more accurate. This issue can be mitigated by collecting or re-sampling more training samples in the low-score area.

\section*{Acknowledgements}
This work was supported in part by the National Natural Science Foundation of China under Project 62272019, and also in part by the International Joint Doctoral Education Fund of Beihang University.

%
%
\bibliographystyle{splncs04}
\bibliography{main}
\end{document}